\documentclass{article}

\PassOptionsToPackage{dvipsnames}{xcolor}

\usepackage{geometry}
\usepackage[round]{natbib}
\usepackage[utf8]{inputenc}
\usepackage[T1]{fontenc}
\usepackage[colorlinks = true,
            linkcolor = blue,
            urlcolor  = blue,
            citecolor = blue]{hyperref}
\usepackage{url}
\usepackage{booktabs}
\usepackage{amsfonts}
\usepackage{nicefrac}
\usepackage{microtype}
\usepackage{amsmath}
\usepackage{amssymb}
\usepackage{amsthm}
\usepackage{algorithm}
\usepackage{algpseudocode}
\usepackage{graphicx}
\usepackage{caption}
\usepackage{subcaption}
\usepackage{relsize}
\usepackage{mathrsfs}
\usepackage{multicol}
\usepackage{wrapfig}
\usepackage{comment}
\usepackage{tikz}
\usetikzlibrary{calc,arrows.meta,decorations.pathreplacing}
\allowdisplaybreaks

\newcommand{\R}{{\mathbb{R}}}
\newcommand{\N}{{\mathbb{N}}}

\usepackage[dvipsnames]{xcolor}

\newtheorem{definition}{Definition}

\title{Edge Flow: A Tractable and Predictive Continuous-Time Model for Gradient Descent at the Edge of Stability}

\author{Pierre Marion \\ Inria, École Normale Supérieure, PSL Research University \\ \texttt{pierre.marion@inria.fr}}

\date{\today}

\begin{document}

\maketitle

\begin{abstract}
Gradient descent in deep learning may operate at the edge of stability (EoS), a regime in which the largest eigenvalue of the loss Hessian hovers near the stability threshold $2/\eta$, where $\eta$ is the learning rate. Classical analysis tools such as gradient flow and the descent lemma do not apply here, motivating the search for a continuous-time model valid at EoS. We propose Edge Flow, a system of three coupled ordinary differential equations that provides a tractable, faithful, and predictive model of gradient descent dynamics at EoS. Edge Flow decomposes the dynamics into a center, an oscillation direction, and an oscillation magnitude. The center follows a modified gradient flow on a symmetrized loss; the direction tracks a top eigenvector of the Hessian via Rayleigh quotient dynamics; and the magnitude grows or decays exponentially depending on whether the sharpness exceeds or falls below the threshold~$2/\eta$. Crucially, sharpness stabilization emerges from the coupled dynamics via a self-stabilization feedback loop. Discretizing Edge Flow only requires two gradient evaluations and one Hessian--vector product at each iteration. We demonstrate empirically that Edge Flow tracks the dynamics of gradient descent at least as faithfully as previously proposed continuous-time EoS models, while in addition resolving the oscillation of the sharpness at the onset of EoS, and that it provides a principled framework for understanding and mitigating instabilities in this regime.
\end{abstract}

\begin{figure}[t]
\centering
\includegraphics[width=\textwidth]{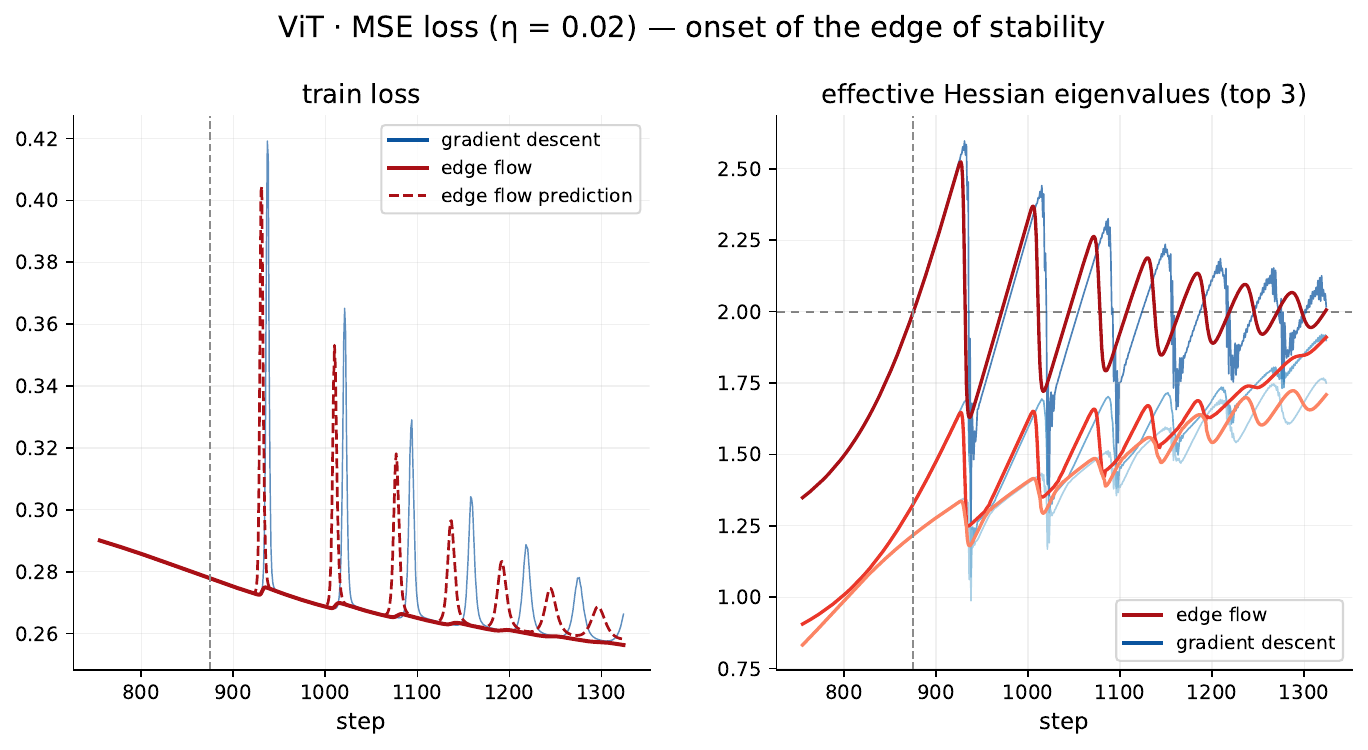}
\caption{\textbf{Edge Flow models gradient descent at the edge of stability} (EoS). Zoom on the onset of EoS for a Vision Transformer trained with mean-squared error loss on a subset of CIFAR-10 ($\eta = 0.02$). \emph{Left:} training loss of gradient descent (blue) and of the Edge Flow (red), with the Edge Flow prediction of the GD loss (dashed red). \emph{Right:} top three eigenvalues of the effective Hessian (i.e.\ eigenvalues multiplied by $\eta$, so that the stability threshold is~$2$). Edge Flow reproduces the loss spikes and sharpness oscillation of gradient descent. The small discrepancy in spikes and oscillations between gradient descent and edge flow is due to discretization errors (see Section \ref{sec:instabilities} for details). Experimental details are given in Appendix~\ref{app:experiments}.}
\label{fig:teaser}
\end{figure}

\section{Introduction}
\label{sec:intro}

Modern deep learning relies on gradient-based optimization with large learning rates. In the influential empirical study of \citet{cohen2021gradient}, building on earlier observations \citep{xing2018walk, jastrzebski2019sharpest, jastrzebski2020breakeven, lewkowycz2020catapult}, it was observed that full-batch gradient descent (GD) on neural network losses consistently operates in a regime called the \emph{edge of stability} (EoS). After an initial phase of \emph{progressive sharpening} during which the largest eigenvalue of the Hessian of the loss $S(w_t) = \lambda_{\max}\!\big(\nabla^2 L(w_t)\big)$ steadily increases, the sharpness $S(w_t)$ reaches the threshold $2/\eta$ and subsequently oscillates around it, neither diverging nor settling. Throughout this process the training loss $L(w_t)$ continues to decrease on average, despite pronounced spikes.

The fine structure of the iterates at EoS is well documented. The GD iterates exhibit \emph{quasi-periodic oscillations}: consecutive iterates $w_t$ and $w_{t+1}$ are displaced from each other primarily along a top eigenvector of the Hessian, with the displacement flipping sign at each step. The result is a characteristic back-and-forth bouncing superimposed on a slow drift.
The mechanism by which the sharpness self-regulates at $2/\eta$ was termed \emph{self-stabilization} by \citet{damian2023selfstabilization}. Informally, when $S(w_t)$ exceeds $2/\eta$, the oscillations grow, which creates a push toward regions of lower curvature; when $S(w_t)$ drops below $2/\eta$, the oscillations damp, the dynamics approach gradient flow, and the progressive sharpening tendency drives the sharpness back up.

Despite this local characterization of the oscillations, understanding global dynamics of GD at the edge of stability is a challenging problem even for convex landscapes, since standard tools from optimization such as the descent lemma do not apply.
A recent body of work has begun analyzing such dynamics in specific settings. These include logistic regression on separable data \citep{wu2023implicit, wu2024large,wu2025large}, simple two-layer networks \citep{ahn2023learning}, diagonal linear networks \citep{even2023sgd}, and losses with normalization or scale invariance \citep{arora2022understanding, lyu2022understanding}. Each of these yields valuable insights, but the conclusions are tied to the particular structure of the loss landscape and do not readily generalize.
A complementary line of work has investigated the stability of the final minimizer \citep{wu2018sgd, mulayoff2021implicit,chen2023edge, mulayoff2024exact,chemnitz2025characterizing,wu2025taking,mulayoff2026stability}. While this question is also crucial, it does not preclude understanding the dynamics of gradient descent along its trajectory. For instance, minima stability does not allow reasoning about implicit regularization by large stepsizes along the trajectory \citep[see, e.g.,][for a practical use case in score matching]{marion2026understanding}.

On the quest for a general understanding of large-stepsize dynamics, a powerful tool would be to derive continuous-time equivalents of the gradient descent iterates. Indeed, the approximation of small-stepsize GD by the gradient flow $\dot w = -\nabla L(w)$ unlocks powerful analytical tools (conserved quantities, implicit bias analyses), which we would like to apply in the large-stepsize regime. By definition, gradient flow is oblivious to the learning rate and cannot capture any phenomenon that depends on the ratio $\eta \cdot S(w)$.

Two proposals have already been made toward a general continuous-time model that is valid at EoS. The first, called \emph{Central Flow}, was introduced by \citet{cohen2025understanding}: it models the slowly drifting center of the oscillating GD iterates by a gradient flow that is projected onto the sublevel set $\{w : S(w) \le 2/\eta\}$. The resulting projection makes the system difficult both to simulate numerically and to analyze theoretically. The second, \emph{Rod Flow} \citep{regis2026rodflow}, models the GD iterates as the endpoints of a rigid rod whose center and orientation evolve continuously. While conceptually appealing, Rod Flow's update rule for the oscillation magnitude does not capture the self-stabilization at the onset of EoS (see Appendix~\ref{app:comparison}).
These shortcomings leave a central open question unresolved:

\begin{center}
\emph{What is a tractable, faithful, and predictive model of GD dynamics at the edge of stability?}
\end{center}

Here, \emph{tractable} means that the model is cheap to simulate and simple enough to be analyzed; \emph{faithful} means that it tracks individual GD trajectories, including the transient regime at the onset of EoS, rather than only reproducing qualitative features; and \emph{predictive} means that it suggests interventions on GD whose effects can be tested empirically.

\paragraph{Contributions.} We propose such a model, which we call \textbf{Edge Flow} (see an example in Figure~\ref{fig:teaser}). Our contributions are the following.

\begin{enumerate}
    \item \textbf{A simple model.} Edge Flow is a system of three coupled ODEs describing the evolution of a center~$\bar w$, an oscillation direction $u \in \mathbb{S}^{d-1}$, and an oscillation magnitude $x > 0$ (Section \ref{sec:model}). Sharpness stabilization at $2/\eta$ arises as a \emph{consequence} of the coupled dynamics, rather than being imposed as a constraint. The model is derived from first principles and elementary observations about the structure of the GD iterates at EoS (Section~\ref{sec:derivation}).

    \item \textbf{A cheap discretization.} Edge Flow admits a natural discretization that we call \emph{Edge Gradient Descent} (EGD). Each iteration of EGD requires two gradient evaluations and one Hessian--vector product, with no eigenvalue decomposition or optimization subroutine.

    \item \textbf{Fine-grained tracking of GD.} Empirically, Edge Flow tracks the trajectory, loss, and sharpness of full-batch GD across architectures and losses (Section~\ref{sec:experiments}). It matches the bulk dynamics while  additionally resolving the overshoot and damped oscillation of the sharpness at the onset of EoS. This exposes the mechanism behind the loss spikes at the onset of EoS, which leads to concrete remedies using EGD: either refining the discretization of the center ODE, or increasing the base level of oscillations to speed up self-stabilization.

\end{enumerate}

\noindent All in all, we believe Edge Flow opens realistic pathways towards a better understanding of dynamics at the edge of stability, and discuss exciting open directions in Section~\ref{sec:open}.

\section{The Edge Flow Model}
\label{sec:model}

This section introduces our continuous-time model for GD at EoS, and its discretization.

\subsection{Setup and notation}
\label{sec:notation}

We consider a loss $L : \R^d \to \R$ that is three times continuously differentiable and denote its gradient by $\nabla L$, its Hessian by $\nabla^2 L$, and its sharpness by $S(w) = \lambda_{\max}(\nabla^2 L(w))$. The learning rate of the gradient descent algorithm is denoted by $\eta > 0$. The unit sphere in $\R^d$ is $\mathbb{S}^{d-1}$, and~$I$ denotes the identity matrix. For a symmetric matrix $A$, we write $\lambda_1(A) \ge \lambda_2(A) \ge \cdots$ for its eigenvalues in nonincreasing order, so that $S(w) = \lambda_1(\nabla^2 L(w))$.

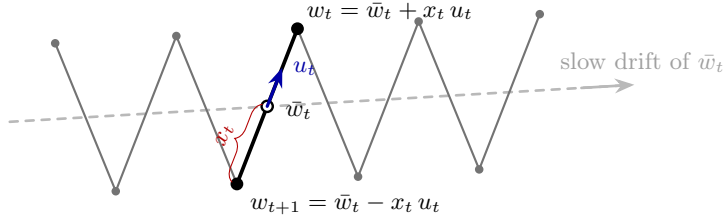
\begin{figure}[t]
\centering
\begin{tikzpicture}[>={Stealth[length=2.6mm]},line cap=round]
  \draw[gray!55,dashed,line width=1pt] (-0.1,-0.006) -- (8.0,0.474);
  \draw[->,gray!55,line width=1pt] (7.45,0.441) -- (8.15,0.483);
  \node[gray!75,anchor=west,font=\small] at (7.05,0.78) {slow drift of $\bar w_t$};
  \draw[black!55,line width=0.8pt]
     (0.5,1.03)--(1.3,-0.922)--(2.1,1.126)--(2.9,-0.826)--(3.7,1.222)--(4.5,-0.73)--(5.3,1.318)--(6.1,-0.634)--(6.9,1.414);
  \foreach \p in {(0.5,1.03),(1.3,-0.922),(2.1,1.126),(4.5,-0.73),(5.3,1.318),(6.1,-0.634),(6.9,1.414)}
     \fill[black!55] \p circle (1.6pt);
  \coordinate (wt)  at (3.7,1.222);
  \coordinate (wt1) at (2.9,-0.826);
  \coordinate (cc)  at (3.3,0.198);
  \draw[black,line width=1.4pt] (wt1) -- (wt);
  \fill (wt) circle (2.4pt);
  \fill (wt1) circle (2.4pt);
  \draw[fill=white,line width=0.9pt] (cc) circle (2.1pt);
  \node[anchor=south west,font=\small] at (wt) {$w_t=\bar w_t+x_t\,u_t$};
  \node[anchor=north west,font=\small] at ($(wt1)+(0.05,-0.02)$) {$w_{t+1}=\bar w_t-x_t\,u_t$};
  \node[anchor=west,font=\small] at ($(cc)+(0.12,-0.06)$) {$\bar w_t$};
  \draw[->,blue!65!black,line width=1.2pt] (cc) -- (3.5,0.71);
  \node[blue!65!black,font=\small,anchor=west] at (3.52,0.70) {$u_t$};
  \draw[decorate,decoration={brace,amplitude=5pt,raise=2pt},red!70!black]
        (wt1) -- (cc)
        node[midway,sloped,above=5pt,red!70!black,font=\small] {$x_t$};
\end{tikzpicture}
\caption{\textbf{Two-timescale structure of gradient descent at EoS, and the Edge Flow notation.} The GD iterates (black) oscillate rapidly back and forth, almost orthogonally to the slow drift of their center $\bar w_t$ (grey): two consecutive iterates straddle $\bar w_t$ and are displaced along a common direction. The decomposition $w_t \approx \bar w_t + (-1)^t x_t u_t$ captures this with three slowly varying quantities: the center $\bar w_t$; its unit direction $u_t$ (blue), which tracks a top Hessian eigenvector; and the oscillation half-magnitude $x_t$ (brace). Edge Flow gives autonomous dynamics for this slow triple $(\bar w_t, u_t, x_t)$.}
\label{fig:rod-schematic}
\end{figure}

The model introduces three dynamical variables (see Figure~\ref{fig:rod-schematic} for an illustration):
\begin{itemize}
    \item the \emph{center} $\bar w_t \in \R^d$, the slowly drifting midpoint of the oscillations;
    \item the \emph{oscillation direction} $u_t \in \mathbb{S}^{d-1}$, a unit vector pointing along the oscillation;
    \item the \emph{oscillation magnitude} $x_t > 0$, the half-length of the oscillation.
\end{itemize}
Following \citet{regis2026rodflow}, the GD iterate at time $t$ is modeled as $w_t \approx \bar w_t + (-1)^t x_t\, u_t$: the iterate alternates between the two endpoints $\bar w + x u$ and $\bar w - x u$ of a segment centered at~$\bar w$. The main insight leading to the model is that the oscillations between two consecutive GD iterates (i.e., between the endpoints) happen at a much faster timescale than the evolution of all other quantities. For this reason, the effective gradient acting on the center of the oscillations is the average of the gradients at the two endpoints.

\subsection{Continuous-time formulation: Edge Flow}
\label{sec:edge_flow_ode}

\begin{definition}[Edge Flow]
\label{def:edge_flow}
Given a learning rate $\eta > 0$, the \emph{Edge Flow} is the system of coupled ODEs, with initial conditions $\bar w_0 \in \R^d$, $u_0 \in \mathbb{S}^{d-1}$, and $x_0 = \varepsilon > 0$:
\begin{align}
    \frac{d\bar w_t}{dt} &= -\,\frac{1}{2}\big(\nabla L \big(\bar w_t + x_t\, u_t\big) + \nabla L \big(\bar w_t - x_t\, u_t\big)\big), \label{eq:ef_w}\\[6pt]
    \frac{du_t}{dt} &= 2 \big(\nabla^2 L(\bar w_t) - \big(u_t^\top \nabla^2 L(\bar w_t)\, u_t\big)\, I\big)\, u_t, \label{eq:ef_u}\\[6pt]
    \frac{dx_t}{dt} &= \big(S(\bar w_t) - \tfrac{2}{\eta}\big)\, x_t. \label{eq:ef_x}
\end{align}
\end{definition}
We make a few preliminary observations about this system.

\paragraph{Well-posedness.} Since $L$ is three times continuously differentiable and $\lambda_{\max}$ is Lipschitz-continuous, the right-hand side of \eqref{eq:ef_w}--\eqref{eq:ef_x} is locally Lipschitz in $(\bar w, u, x)$, so the system admits a unique maximal solution by the Picard--Lindel\"of theorem \citep{hartman2002ordinary}. Moreover, the sphere $\{\|u\| = 1\}$ and the half-line $\{x > 0\}$ are invariant under the dynamics (see below for the former), so the three variables retain their interpretation for all times.

\paragraph{Structure of the three equations.} Equation~\eqref{eq:ef_w} drives the center by a gradient flow on the \emph{symmetrized loss}---the average of $L$ evaluated at the two endpoints. This is the same equation as in \citet{regis2026rodflow}. Importantly, the two other equations significantly depart from Rod Flow (see Appendix~\ref{app:comparison} for details). Equation~\eqref{eq:ef_u} is the \emph{Rayleigh quotient gradient flow} on the unit sphere, closely related to Oja's flow \citep{oja1982simplified}: it drives $u$ toward the top eigenvector of the Hessian at $\bar{w}_t$. Equation~\eqref{eq:ef_x} is a scalar linear ODE with time-varying coefficient $S(\bar w_t) - 2/\eta$: the oscillation magnitude grows exponentially when $S > 2/\eta$ and decays when $S < 2/\eta$. The learning rate~$\eta$ enters the system only through the sharpness threshold in~\eqref{eq:ef_x}.

\paragraph{Recovering gradient flow.} Initializing $x_0 = 0$ makes $x_t = 0$ for all $t$. In that case \eqref{eq:ef_w} reduces to the standard gradient flow. Edge Flow can therefore be viewed as a \emph{deformation} of gradient flow by the oscillation $(x, u)$, which activates only when the sharpness reaches $2/\eta$.

\paragraph{Self-stabilization.} The three equations are coupled through the center $\bar w$: equation~\eqref{eq:ef_x} reads off the sharpness at the center, while equation~\eqref{eq:ef_w} depends on the endpoints $\bar w \pm xu$. When $S(\bar w) > 2/\eta$, the magnitude $x$ grows; a larger $x$ causes the symmetrized gradient in \eqref{eq:ef_w} to depart from $\nabla L(\bar w)$ and steer $\bar w$ toward regions of lower curvature (see Section~\ref{sec:derive_w} for details), which in turn decreases $S(\bar w)$. Conversely, when $S(\bar w) < 2/\eta$, the magnitude $x$ shrinks, the symmetrized gradient approaches $\nabla L(\bar w)$, and the progressive-sharpening tendency of gradient flow pushes the sharpness back up. The pair $(S(\bar w), x)$ thus behaves like a negative-feedback oscillator attracting $S(\bar w)$ to $2/\eta$. This reproduces the self-stabilization mechanism identified by \citet{damian2023selfstabilization} as a \emph{consequence} of the dynamics rather than an explicit constraint.

\paragraph{The base level $\varepsilon$.} The parameter $\varepsilon > 0$ is the initial oscillation magnitude. Physically, it represents the small perturbations---due to numerical errors or inherent parameter jitter---that seed the instability once the sharpness reaches $2/\eta$. 
A larger $\varepsilon$ leads to faster activation of the self-stabilization mechanism, as the exponential growth in~\eqref{eq:ef_x} starts from a larger base. 
We refer to Section~\ref{sec:experiments} for more details on the numerical implementation of \eqref{eq:ef_x}.

\paragraph{About Equation~\eqref{eq:ef_u}: the Rayleigh quotient flow.} The ODE \eqref{eq:ef_u} is the gradient ascent flow for the Rayleigh quotient $R(u) = u^\top \nabla^2 L(\bar w)\, u$ restricted to the unit sphere $\mathbb{S}^{d-1}$. Its Riemannian gradient is $2(I - uu^\top)\nabla^2 L(\bar w)\, u$, which is exactly the right-hand side of~\eqref{eq:ef_u}. One can verify that this flow preserves the unit norm: $\frac{d}{dt}\|u\|^2 = 0$ whenever $\|u\|=1$. When $\nabla^2 L(\bar w)$ has a simple largest eigenvalue $\lambda_1 > \lambda_2$, the corresponding eigenvector is an attracting fixed point, with exponential convergence rate governed by the spectral gap $\lambda_1 - \lambda_2$ \citep{helmke1994optimization}. When $\nabla^2 L(\bar w)$ changes slowly with time, $u_t$ tracks its top eigenvector.

\subsection{Discrete-time formulation: Edge Gradient Descent}
\label{sec:egd}

Edge Flow has a natural discretization obtained by applying the forward Euler method to \eqref{eq:ef_w} and~\eqref{eq:ef_x}, and the power method to \eqref{eq:ef_u}. The result is an iterative algorithm that we call \emph{Edge Gradient Descent} (EGD). This algorithm gives a way to efficiently simulate Edge Flow, which is useful to analyze the behavior of large-stepsize GD, and is also interesting on its own, as it seems to be numerically more stable than GD by taking a discretization step $\rho < \eta$.

\begin{definition}[Edge Gradient Descent]
\label{def:egd}
Given a learning rate $\eta > 0$ and a discretization step $\rho > 0$ with $\rho \le \eta$, \emph{Edge Gradient Descent} is defined, for $k \in \N$, by
\begin{align}
    \bar w_{k+1} &= \bar w_k - \frac{\rho}{2}\Big(\nabla L \big(\bar w_k + x_k\, u_k\big) + \nabla L \big(\bar w_k - x_k\, u_k\big)\Big), \label{eq:egd_w}\\[6pt]
    u_{k+1} &= \frac{\nabla^2 L(\bar w_k)\, u_k}{\big\|\nabla^2 L(\bar w_k)\, u_k\big\|}, \label{eq:egd_u}\\[6pt]
    x_{k+1} &= x_k \Big(1 + \rho\Big(S(\bar w_k) - \tfrac{2}{\eta}\Big)\Big), \label{eq:egd_x}
\end{align}
initialized at $\bar w_0 = w_0$, $u_0 \in \mathbb{S}^{d-1}$, and $x_0 = \varepsilon > 0$.
\end{definition}

\paragraph{Two distinct stepsizes.} The stepsizes $\rho$ and $\eta$ play fundamentally different roles. The learning rate $\eta$ is the stepsize of the original GD algorithm; it enters through the sharpness threshold $2/\eta$ in~\eqref{eq:egd_x}. The parameter $\rho$ is the integration step for the Edge Flow ODEs and controls the discretization accuracy. In short, $\eta$ is part of the \emph{model}, while $\rho$ is part of the \emph{numerical method}. In particular, taking $\rho < \eta$ does not alter the sharpness threshold $2/\eta$, while resolving the center dynamics more finely, which is important in stiff regions (see Section~\ref{sec:instabilities}).

\paragraph{Computational cost.} Each EGD iteration requires two gradient evaluations $\nabla L(\bar w_k \pm x_k u_k)$ and one Hessian--vector product $\nabla^2 L(\bar w_k)\, u_k$, computable at a cost comparable to one gradient evaluation via automatic differentiation.
The sharpness $S(\bar w_k)$ is obtained at no extra cost as the norm $\|\nabla^2 L(\bar w_k)\, u_k\|$. The time variable of Edge Flow is calibrated so that one GD step corresponds to a time interval $\eta$ of the flow (see Section~\ref{sec:derivation}); simulating one GD step therefore requires $K := \eta/\rho$ EGD iterations. 
Notably, at the coarsest setting $\rho = \eta$, EGD is thus about three times as expensive as GD. 
By contrast, each step of Central Flow requires extracting top eigenpairs of the Hessian and solving a semidefinite complementarity problem for the oscillation covariance; Rod Flow propagates a low-rank orientation matrix~$\Sigma$ which involves at each step a Gram-Schmidt orthogonalization and a QR decomposition.

\section{Derivation from First Principles}
\label{sec:derivation}

We now derive each of the three Edge Flow equations from elementary observations about the structure of GD at the edge of stability. We emphasize that our goal is \emph{not} to construct a continuous-time dynamics that exactly interpolates the GD iterates, but rather to find the \emph{simplest} system that locally tracks GD and reproduces its key global patterns.

We also note that the derivation below neglects higher-order coupled corrections between $\bar w$, $u$, and $x$. We expect these to contribute at higher order in $\eta$ relative to the leading terms. Our aim in this paper is to provide an intuitive derivation of the model, but we do not claim that our derivation universally applies  (for instance, it would not apply to quadratic objectives). Deriving precise conditions under which it applies is an interesting open question (Section~\ref{sec:open}).

\subsection{Two-timescale decomposition}
\label{sec:twotimescale}

The starting observation is that the GD iterates at EoS exhibit a \emph{separation of timescales}. At each step, the iterate makes a large displacement along the top eigenvector of the Hessian, with the sign of this displacement alternating. Over two consecutive steps, the large displacements nearly cancel, leaving only a small net movement. This motivates decomposing the iterate as
\begin{equation}\label{eq:decomp}
    w_t \;\approx\; \bar w_t + (-1)^t\, x_t\, u_t,
\end{equation}
where $\bar w_t$ is the slowly drifting center, $u_t$ is the direction of oscillation (a unit vector), and $x_t > 0$ is the oscillation half-amplitude. The three components $(\bar w_t, u_t, x_t)$ all change slowly from one step to the next, while the sign $(-1)^t$ encodes the fast oscillation. The geometry of the decomposition and the associated notation are summarized in Figure~\ref{fig:rod-schematic}.

The key empirical fact that justifies this decomposition is that \emph{consecutive gradients are nearly anti-aligned} at EoS. Indeed, a Taylor expansion gives
\begin{align*}
    \nabla L(\bar w + x u) \approx \nabla L(\bar w) + x\,\nabla^2 L(\bar w)\, u, \quad
    \nabla L(\bar w - x u) \approx \nabla L(\bar w) - x\,\nabla^2 L(\bar w)\, u.
\end{align*}
When the oscillation amplitude $x$ is large enough that $\|x\,\nabla^2 L(\bar w)\, u\| \gg \|\nabla L(\bar w)\|$, the two gradients point in nearly opposite directions. 
This near-anti-alignment of the gradients at consecutive iterates at EoS has been experimentally documented across architectures in prior work \citep{cohen2021gradient, damian2023selfstabilization, cohen2025understanding}.

\subsection{Dynamics of the center}
\label{sec:derive_w}

To derive the evolution of $\bar w$, we compute the evolution of the center after one gradient step, as in Rod Flow \citep{regis2026rodflow},
\begin{align*}
    \bar{w}_{t+1} - \bar{w}_t &= \frac{w_{t+2} + w_{t+1}}{2} - \frac{w_{t+1} + w_{t}}{2} \\
    &= \frac{w_{t+2} - w_{t+1}}{2} + \frac{w_{t+1} - w_{t}}{2} \\
    &= - \frac{\eta}{2}\big(\nabla L(w_{t+1}) + \nabla L(w_t)\big) \\
    &\approx  - \frac{\eta}{2}\big(\nabla L\!\big(\bar w_t + x_t\, u_t\big) + \nabla L\!\big(\bar w_t - x_t\, u_t\big)\big) ,
\end{align*}
where the last equality uses the decomposition~\eqref{eq:decomp} together with the slow variation of $(\bar w_t, u_t, x_t)$, so that $w_{t+1} \approx \bar w_t - x_t u_t$ up to higher-order corrections.
We recover exactly the update rule from \eqref{eq:egd_w} when $\rho = \eta$. The main assumption of Edge Flow is that this update rule is relatively insensitive to the stepsize. Intuitively, the reason is that the stepsize-sensitive component of the gradients---the component along $u_t$, which sees a curvature close to $2/\eta$---cancels in the sum. The symmetrized gradient is therefore supported, to leading order, on directions in which the curvature is less than $2/\eta$. In those directions gradient updates are stable and well approximated by a flow for any stepsize $\rho \le \eta$. Letting $\rho \to 0$ yields
\[
\frac{d\bar w_t}{dt} = -\,\frac{1}{2}\big(\nabla L\!\big(\bar w_t + x_t\, u_t\big) + \nabla L\!\big(\bar w_t - x_t\, u_t\big)\big),
\]
which is equation~\eqref{eq:ef_w}. This is a gradient flow on the \emph{symmetrized loss}: by a Taylor expansion,
\begin{equation}\label{eq:symgrad}
    \frac{1}{2}\Big(\nabla L(\bar w + xu) + \nabla L(\bar w - xu)\Big) = \nabla L(\bar w) + \frac{x^2}{2}\,\nabla^3 L(\bar w)[u, u] + O(x^4).
\end{equation}
At the first order, the center dynamics thus differ from pure gradient flow by a third-derivative correction that is quadratic in the oscillation amplitude. This correction is the mechanism behind the self-stabilization at EoS already identified by \citet{damian2023selfstabilization}: it biases the center toward regions of parameter space where the curvature in the direction $u$ is smaller.

\subsection{Dynamics of the oscillation direction}
\label{sec:derive_u}

The oscillation direction $u_t$ should align with one of the dominant eigenvectors of the Hessian $\nabla^2 L(\bar w_t)$, since this is the direction in which the GD iterates are most unstable. Empirically, the direction $u_t$ evolves slowly, even when multiple eigenvalues are near $2/\eta$: because consecutive gradients are nearly anti-aligned, the net rotation of the oscillation axis per step is small.

We model this slow evolution by the \emph{power method}, the classical algorithm for computing the dominant eigenvector of a matrix \citep{golub2013matrix}. In discrete time, one power iteration applied to the Hessian $\nabla^2 L(\bar w_t)$ with current direction $u_t$ yields
\begin{equation*}
    u_{t+1} = \frac{\nabla^2 L(\bar w_t)\, u_t}{\|\nabla^2 L(\bar w_t)\, u_t\|},
\end{equation*}
which is equation~\eqref{eq:egd_u}. This choice is motivated by the observation that GD itself performs implicit power iterations: linearizing the GD update around the center, the oscillation component is multiplied at each step by $(I - \eta\,\nabla^2 L(\bar w))$. Repeated application of this matrix amplifies the component along the top eigenvector, exactly as the power method does.

The continuous-time limit of the power method is the Rayleigh quotient gradient ascent flow on the unit sphere \citep{oja1982simplified, helmke1994optimization}:
\begin{equation*}
    \frac{du}{dt} = 2 \big(\nabla^2 L(\bar w) - u^\top \nabla^2 L(\bar w)\, u\, I \big) \, u,
\end{equation*}
which is equation~\eqref{eq:ef_u}.
We note that the precise time normalization of \eqref{eq:ef_u} (for instance, the factor $2$) does not matter much for the model: all that matters is that the direction dynamics are fast compared with the drift of the center, so that $u_t$ tracks the top eigenvector adiabatically.

When multiple eigenvalues of the Hessian are simultaneously near $2/\eta$, the spectral gap is small, the power method converges slowly, and the oscillation is genuinely spread across a multi-dimensional top eigenspace. This corresponds to the empirical observation that the direction of the oscillation varies relatively quickly \citep{cohen2025understanding}. Central Flow thus models the oscillation direction as a random variable whose covariance matrix across the top eigenspace can be tracked, however at the cost of solving a semidefinite complementarity problem. We instead make the oscillation direction evolve deterministically as part of the coupled dynamics.

\subsection{Dynamics of the oscillation magnitude}
\label{sec:derive_x}

We now derive the equation governing the oscillation half-amplitude $x_t$.
To this aim, starting from $w_t = \bar w_t + x_t u_t$, we expand one GD step to first order in the oscillation magnitude:
\begin{align*}
    w_{t+1} &= w_t - \eta \nabla L(w_t) \\
    &= \bar w_t + x_t u_t - \eta \nabla L(\bar w_t) - \eta x_t \nabla^2 L(\bar w_t)\, u_t + \mathcal{O}(x_t^2) \\
    &= \bar w_t - \eta \nabla L(\bar w_t) + x_t (I - \eta \nabla^2 L(\bar w_t)) u_t + \mathcal{O}(x_t^2) \\
    &= \bar w_{t+1} + x_t (I - \eta \nabla^2 L(\bar w_t)) u_t + \mathcal{O}(x_t^2),
\end{align*}
where the last equality uses that $\bar w_{t+1} = \bar w_t - \eta \nabla L(\bar w_t) + \mathcal{O}(x_t^2)$, by \eqref{eq:symgrad}.
Since $u_t$ is close to the top eigenvector, we have $(I - \eta \nabla^2 L(\bar w_t))\, u_t \approx (1 - \eta S(\bar w_t))\, u_t$, so the signed oscillation magnitude after one step is approximately $x_t(1 - \eta S(\bar w_t))$. Since $\eta S(\bar w_t) \approx 2$ at EoS, this factor is close to $-1$: the sign flips at each step, confirming the period-two oscillatory behavior. Absorbing the sign flip into the factor $(-1)^t$ of the decomposition~\eqref{eq:decomp}, the unsigned magnitude evolves as
\[
x_{t+1} = x_t (\eta S(\bar w_t) - 1) = x_t \Big(1 + \eta \big(S(\bar w_t) - \frac{2}{\eta}\big)\Big) \, .
\]
Note that this equation was already obtained by \citet{damian2023selfstabilization}, but was not used to derive the subsequent Central Flow \citep{cohen2025understanding}.
Similarly to the dynamics of the center, we assume that this update rule is relatively insensitive to the update rate $\eta$ because the change in $S(\bar w_t)$ happens slowly compared to the fast oscillations. For this reason, we can replace the outer $\eta$ by a smaller stepsize $\rho < \eta$, giving 
$
x_{k+1} = x_k (1 + \rho (S(\bar w_k) - \frac{2}{\eta})) ,
$
which is equation \eqref{eq:egd_x}. Taking $\rho \to 0$ gives equation~\eqref{eq:ef_x}. The oscillation magnitude grows exponentially when $S > 2/\eta$ and decays when $S < 2/\eta$, with the equilibrium occurring at the EoS threshold.

\section{Experiments}
\label{sec:experiments}

We first describe the experimental setup and the predictions extracted from Edge Flow (Section~\ref{sec:exp-setup}), then show that it tracks gradient descent and reproduces self-stabilization (Section~\ref{sec:exp-tracking}), and finally discuss instabilities at the onset of the edge of stability (Section~\ref{sec:instabilities}).

\subsection{Setup and discretization scheme}
\label{sec:exp-setup}

\paragraph{Protocol.} Following \citet{cohen2025understanding}, we validate Edge Flow on several configurations, spanning four architectures (a multi-layer perceptron, a convolutional network, a residual network, and a vision transformer) and two losses (mean-squared error and cross-entropy), each trained on a small subset of CIFAR-10. For every configuration we run full-batch GD and Edge Gradient Descent from the same initialization; for the CNN with MSE loss we additionally run the Central Flow of \citet{cohen2025understanding}. Our baseline is Central Flow, because it is to our knowledge the only alternative flow able to stabilize at the onset of EoS. We refer to Appendix~\ref{app:comparison} for discussion of Rod Flow. Full experimental details are collected in Appendix~\ref{app:experiments}.

\paragraph{Discretization.} The flows are integrated with forward Euler, splitting each modeled GD step of size $\eta$ into $K = \eta/\rho$ substeps (Section~\ref{sec:egd}). For Edge Flow, this exactly corresponds to using the proposed Edge Gradient Descent. Unless stated otherwise we use the default value of $K=4$ for Central Flow and $K=8$ for Edge Flow (we checked that $K=4$ gives similar results). The discretization matters chiefly in stiff regions, in particular at the onset of EoS (Section~\ref{sec:instabilities}). We leave more efficient discretization schemes to later study.

The center is initialized at the GD initialization $\bar w_0 = w_0$, the oscillation direction $u_0$ at the top eigenvector of the Hessian, and the oscillation magnitude at a small value seeding the instability. Specifically, to prevent $x_t$ from exponential decay to machine-error precision during the initial progressive sharpening phase where $S < 2/\eta$, we impose a floor on the magnitude, $x_t \leftarrow \max(x_t, \varepsilon)$. The value $\varepsilon$ is either set as a hyperparameter (see Section~\ref{sec:instabilities}) or taken as $\varepsilon_t := |\nabla L(\bar{w}_t)^\top u_t|$ to remove a hyperparameter and track gradient descent in Section~\ref{sec:exp-tracking}.

\begin{figure}[t!]
\centering
\includegraphics[width=0.9\textwidth]{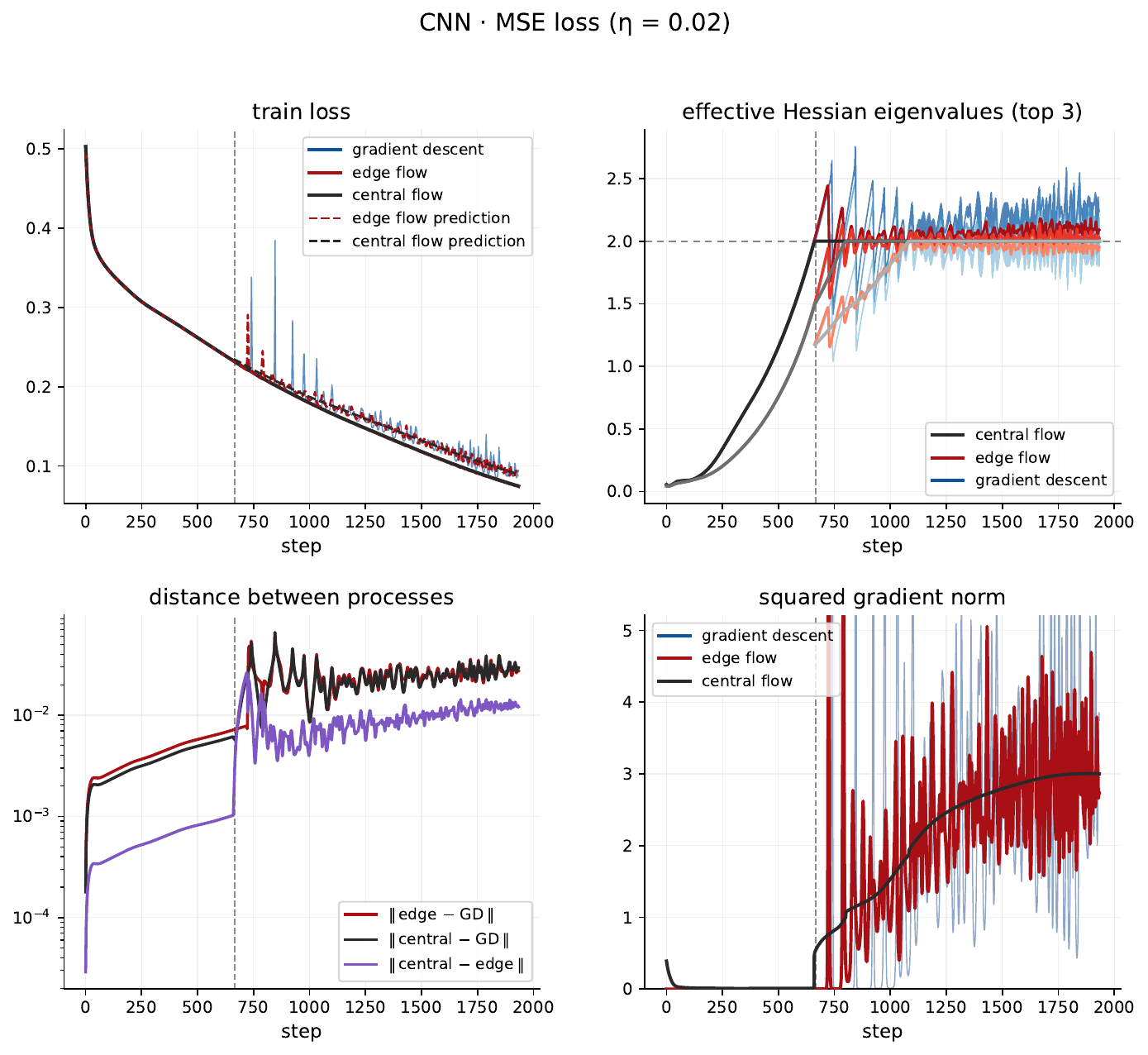}
\caption{
    \textbf{Gradient descent, Edge Flow, and Central Flow for a CNN with MSE loss} on a $1000$-example, $4$-class subset of CIFAR-10 ($\eta = 0.02$). Throughout, gradient descent is shown in blue, Edge Flow in red, and Central Flow \citep{cohen2025understanding} in black; the vertical dashed line marks the onset of the edge of stability. \emph{Top left:} training loss, with each flow's prediction of the GD loss (dashed, see main text for details on the Edge Flow prediction). The Edge Flow and Central Flow curves are superposed. Both flows make good predictions for the GD train loss. \emph{Top right:} the top three eigenvalues of the effective Hessian (eigenvalue $\times\,\eta$, so the threshold is~$2$). Within a colour, the shade encodes the eigenvalue rank. Both flows track the bulk dynamics, and Edge Flow is also able to capture the oscillatory dynamics around $2/\eta$ (see zoom in Figure~\ref{fig:cnn-mse-eos}). \emph{Bottom left:} the pairwise distances in parameter space between the three processes (log scale). The two flows track GD and are close to one another. \emph{Bottom right:} the squared gradient norm (true value for GD, predictions for the flows, see main text for details on the Edge Flow prediction). Edge Flow and Central Flow both make reasonable predictions for the evolution of the squared gradient norm. Central Flow predicts the moving average while Edge Flow reproduces the oscillatory nature of the norm. 
    }
\label{fig:cnn-mse-overlay}
\end{figure}

\paragraph{Edge Flow predictions.}
Edge Flow models the GD iterates as oscillating between the two endpoints $\bar w_t \pm x_t u_t$ (Section~\ref{sec:model}). Similar to \citet{cohen2025understanding}, we can deduce predictions for key quantities along the GD iterates:
\begin{itemize}
    \item \textbf{Loss.} The predicted loss is the mean of the loss at the two endpoints, $\widehat L_t := \tfrac12\big(L(\bar w_t + x_t u_t) + L(\bar w_t - x_t u_t)\big)$. A second-order expansion gives $\widehat L_t \approx L(\bar w_t) + \tfrac12 x_t^2\, u_t^\top \nabla^2 L(\bar w_t)\, u_t = L(\bar w_t) + \tfrac12 x_t^2\, S(\bar w_t)$, so at the first order the averaged loss exceeds the loss at the center by half the curvature times the oscillation variance.
    \item \textbf{Squared gradient norm.} Writing $g_\pm = \nabla L(\bar w_t \pm x_t u_t)$ for the two endpoint gradients, the time-averaged squared gradient norm splits into a center term and an oscillation term, $\tfrac12\big(\|g_+\|^2 + \|g_-\|^2\big) = \tfrac14\big\|g_+ + g_-\big\|^2 + \tfrac14\|g_+ - g_-\|^2$. At the edge of stability the gradient norm is dominated by the oscillation term $\tfrac14\|g_+ - g_-\|^2$, which is the quantity we report.
\end{itemize}

\noindent We can also make a prediction for the oscillation variance, see details in Appendix~\ref{app:experiments}. 

\subsection{Edge Flow tracks GD dynamics and implements self-stabilization}
\label{sec:exp-tracking}

\begin{figure}[t]
\centering
\includegraphics[width=0.9\textwidth]{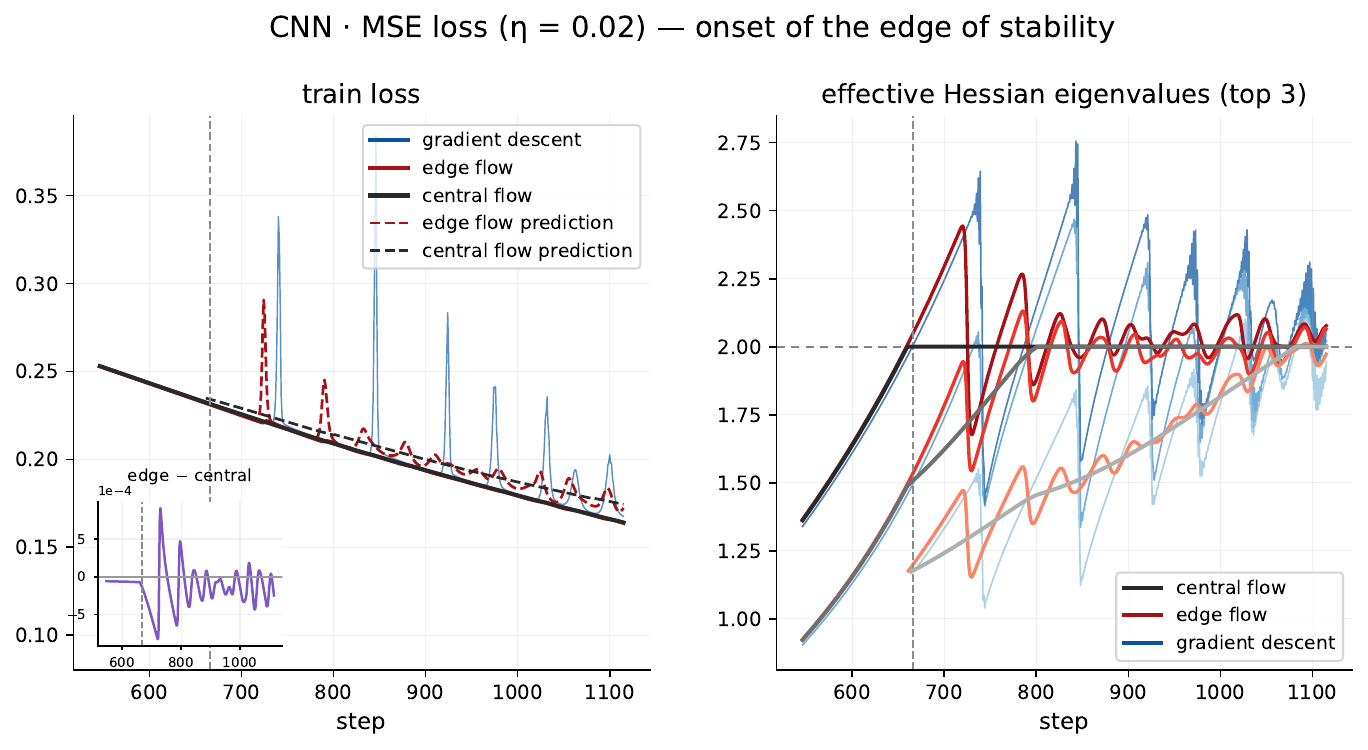}
\caption{\textbf{Onset of the edge of stability for the CNN with MSE loss} (zoom of Figure~\ref{fig:cnn-mse-overlay}). \emph{Left:} training loss; the inset shows the small gap between the Edge Flow and Central Flow centers. \emph{Right:} the top three effective-Hessian eigenvalues reach the threshold~$2$ in staggered fashion and then oscillate around it with damping. Edge Flow reproduces the overshoot and ringing of gradient descent, which the instantaneous projection of Central Flow does not.}
\label{fig:cnn-mse-eos}
\end{figure}

Figure~\ref{fig:cnn-mse-overlay} compares full-batch gradient descent with Edge Flow and Central Flow on a convolutional network trained with MSE loss. The analogous experiments for other architectures and losses (with Edge Flow only), as well as additional metrics, are deferred to Appendix~\ref{app:additional-figures}. Across every metric, the two flows match the GD trajectory. For the train loss (top left plot of Figure~\ref{fig:cnn-mse-overlay}), the Edge Flow prediction tracks the GD iterates (see previous section for details). Interestingly, in weight space, the two flows stay closer to each other than either is to gradient descent (bottom left plot of Figure~\ref{fig:cnn-mse-overlay}).

The two models agree quantitatively in Figure~\ref{fig:cnn-mse-overlay} but differ in how the oscillation is obtained. Central Flow \citep{cohen2025understanding}
models $w_t = w(t) + \delta_t$ with $\mathbb{E}[\delta_t]=0$ and $\mathbb{E}[\delta_t \delta_t^\top] = \Sigma_t$, and determines $\Sigma$ at every step by solving a stationarity (semidefinite complementarity) condition, so the amplitude is \emph{instantaneously} controlled by the current sharpness. Edge Flow instead evolves the magnitude $x_t$ through its own differential equation~\eqref{eq:ef_x}, so the amplitude \emph{lags} behind the sharpness. This delay is precisely what produces the overshoot and damped oscillation of the sharpness at the onset of the edge of stability seen in Figure~\ref{fig:cnn-mse-eos}. 
Edge Flow reproduces this dynamics of self-stabilization, whereas Central Flow's projection clamps the sharpness to $2/\eta$. We observe some discrepancies in the Edge Flow oscillations and the GD ones, see the next section for a discussion.

Once several eigenvalues reach the edge, the GD sharpness settles somewhat \emph{above} $2/\eta$. Interestingly, Edge Flow exhibits a similar (albeit smaller) residual. However, we hypothesize that this has no appreciable bearing on the loss trajectory or the implicit regularization.

\subsection{Instabilities at the onset of Edge of Stability}
\label{sec:instabilities}

\begin{figure}[t]
\centering
\includegraphics[width=\textwidth]{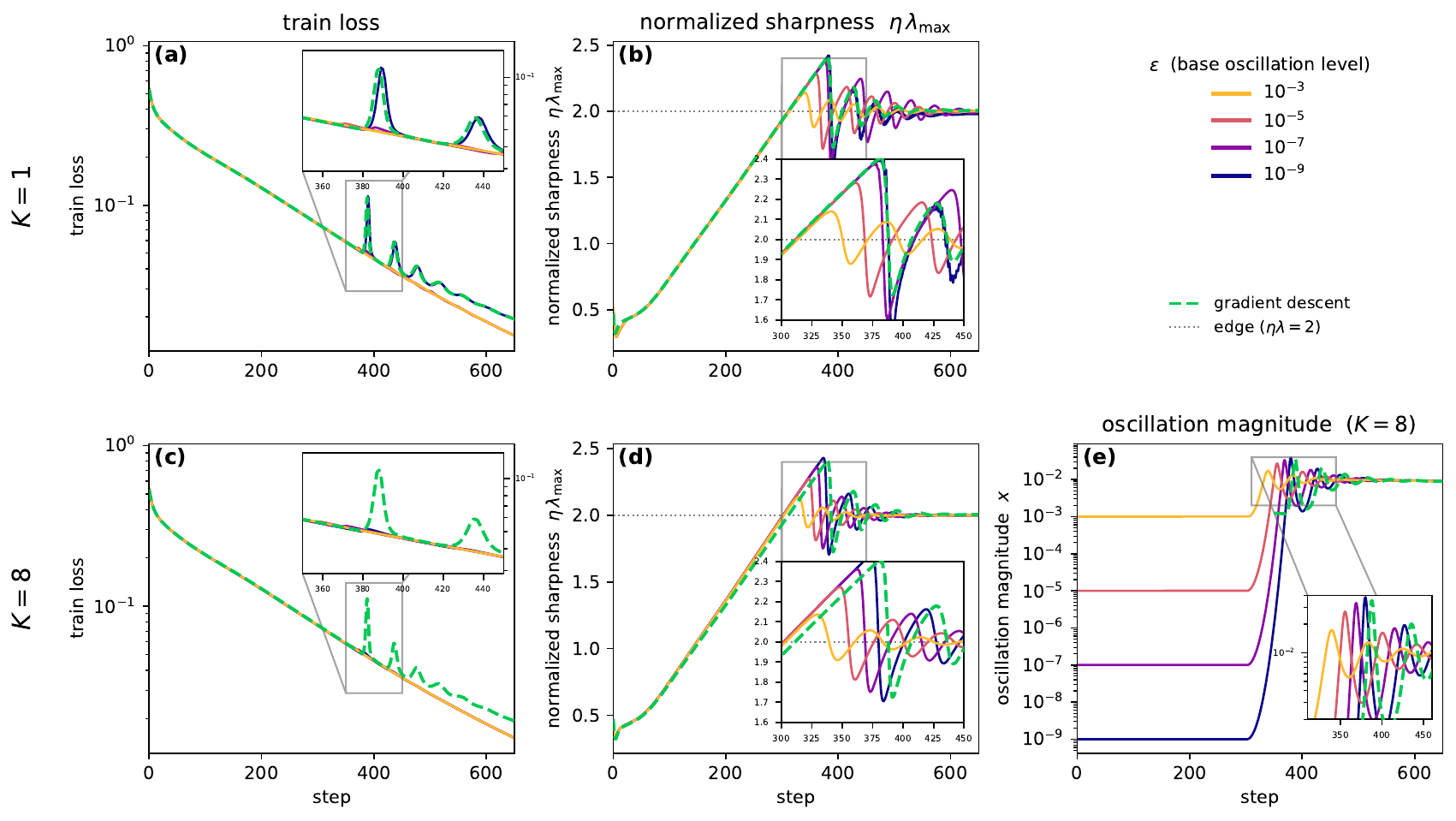}
\caption{ 
\textbf{Effect of the base oscillation level $\varepsilon$ and of the discretization $K$ on the EoS onset}, for a $3$-layer MLP trained with MSE loss on a $400$-example, $4$-class subset of CIFAR-10 ($\eta = 0.02$). \emph{Rows:} a coarse ($K=1$) and a fine ($K=8$) discretization of Edge Gradient Descent ($\rho = \eta/K$). \emph{Columns:} training loss, normalized sharpness $\eta\lambda_{\max}$, and---for $K=8$---the oscillation magnitude $x$. As $\varepsilon$ increases, the sharpness overshoot and the loss spike both shrink; the post-onset dynamics are unchanged, the magnitude $x$ converging to the same plateau (the GD oscillation amplitude) regardless of $\varepsilon$. A large value of $K$ suppresses the loss spikes without affecting the self-stabilization dynamics.}
\label{fig:onset-base-level}
\end{figure}

The delay in the response of $x_t$ discussed above leads to another interesting phenomenon: when $x_t$ increases exponentially fast to non-negligible values, the second order term in \eqref{eq:symgrad} kicks in, which makes the gradient take a stiff turn in parameter space. 
This produces a transient \emph{loss spike}.
Interestingly, the Edge Flow model exposes two handles on this instability, which we examine on a small 3-layer MLP (Figure~\ref{fig:onset-base-level}), leaving larger scale experiments to future work.

\paragraph{Refining the discretization of the center.} Decreasing the center step $\rho = \eta/K$ by increasing $K$ integrates the stiff turn of equation~\eqref{eq:ef_w} more accurately and suppresses the loss spike, without altering the self-stabilization itself (left panels of Figure~\ref{fig:onset-base-level}).
In other words, our model suggests that the loss spike is a discretization artifact rather than an intrinsic feature of self-stabilization. A finer  discretization also seems to improve stability: it remains stable at learning rates $\eta$ at which the coarse $K=1$ scheme (and gradient descent itself) diverge (see Appendix~\ref{app:onset}). This suggests that adaptive stepsize refinement near the onset of EoS---standard in the stiff-ODE literature \citep{hairer1996solving}---could mitigate loss spikes and catastrophic divergence, with the cost of the extra gradient evaluations incurred only when needed.

\paragraph{Increasing the base level of oscillations.}
The floor $\varepsilon$ on the oscillation magnitude $x_t$ is a second way to tune the delay. A larger $\varepsilon$ lets the exponential growth of $x_t$ engage sooner once the sharpness crosses $2/\eta$, so self-stabilization pulls the sharpness back more promptly. Figure~\ref{fig:onset-base-level} confirms this: as $\varepsilon$ increases, both the sharpness overshoot and the onset loss spike decrease; the effect is consistent across the coarse ($K=1$) and fine ($K=8$) discretizations. In large-scale training, this suggests that injecting a small amount of noise into the iterates along the top-eigenvector direction should improve stability at the onset of EoS. 

\medskip

\noindent We also observe that the frequency of oscillations is sensitive to the choice of $\varepsilon$ and $K$, the best fit to GD being with small $\varepsilon$ and $K=1$. This explains the discrepancy observed in Figure~\ref{fig:teaser} and other figures where we took $K=8$. Intuitively, this is due to the stiff dynamics in $x_t$: a small discretization error leads to a large offset for the oscillation. We believe that these discrepancies do not matter much for the overall learning dynamics and implicit regularization effect of large stepsizes.

\section{Discussion}
\label{sec:open}

Edge Flow provides a continuous-time model of gradient descent at the edge of stability that is simple enough to simulate, analyze, and build upon. We outline several directions that become tractable now that such a model is available.

\paragraph{Extending the Edge Flow model.}
The present model concerns full-batch gradient descent. Extending it to stochastic gradient descent and adaptive methods is a natural next step. Another axis is to do a rigorous error analysis between GD iterates and Edge Flow, as well as rigorously proving the self-stabilization feedback loop of the Edge Flow iterates.

\paragraph{Applying Edge Flow to theoretical questions.}
Can Edge Flow be used to prove implicit regularization results for large learning rates in tractable settings? This would advance our understanding of the interplay of loss curvature with architecture, features, and generalization.

Does Edge Gradient Descent converge provably faster than standard GD? The decoupling of the two stepsizes $\rho$ and $\eta$ offers new degrees of freedom: one can take aggressive steps for the sharpness control while using a fine discretization for the center. In logistic regression, large learning rates are known to accelerate convergence to max-margin classifiers \citep{wu2024large,wu2025large}; Edge Flow may provide the right framework for isolating the mechanism of this speed-up and transferring it to richer settings.

\paragraph{Methodological improvements.}
Can one design discretization schemes for Edge Flow that are more efficient than forward Euler? Schemes tailored to the stiff nature of the equations could reduce the cost while maintaining accuracy. A particularly intriguing possibility is a discretization that is faster than GD itself, by leveraging larger steps in the directions orthogonal to the oscillation, where the dynamics are smooth.
Finally, the analysis of instabilities in Section~\ref{sec:instabilities} suggests concrete strategies for mitigating a source of loss spikes in large-scale training.

\section*{Acknowledgments}

This paper has benefited from discussions with Lénaïc Chizat, Jeremy Cohen, Alex Damian, Léo Dana, Antoine Gonon, Rotem Mulayoff, Fabian Schaipp, and Aditya Varre. The author is especially grateful to Nicolas Boumal for having reminded them of the Rod Flow construction in a timely manner, to Ananya Harsh Jha for discussions on experimental results at edge of stability, and to Francis Bach for careful proofreading and suggestions. 

\section*{Disclosure of funding, LLM usage, climate impact}

This research has benefited from funding by Google, Meta, and the Prairie French AI Cluster. The author used Claude Opus 4.8 (Anthropic) as a coding and writing assistant, to help write boilerplate experimental code, generate the figures, cross-check the manuscript against the implementation, and edit and proofread the text; all results, claims, and final wording were verified by the author. Using \citet{mistral2025environmental}, we very roughly estimate at 10 kg CO2e the associated CO2 emissions (400k tokens, and multiplying by 10 the emissions per token for Claude Opus 4.8 compared to Mistral Large~2). The CO2 emissions caused by the experiments are estimated below 10 kg CO2e (the bulk of which comes from running a RTX Pro 6000 for around 15 hours on the USA power grid).

\bibliographystyle{abbrvnat}
\bibliography{references_revised}

\appendix

\section{Comparison with Rod Flow}
\label{app:comparison}

Two previous continuous-time models have been proposed for gradient descent at the edge of stability: Central Flow \citep{cohen2025understanding} and Rod Flow \citep{regis2026rodflow}.
These two models as well as Edge Flow share the same starting point: at EoS, the GD trajectory splits into a slowly moving center and a fast oscillation, and one seeks autonomous dynamics for the slow variables. They differ in two essential choices. The first is how the oscillation is \emph{represented}: a constraint on the sharpness for Central Flow, a rod with an orientation matrix $\Sigma$ for Rod Flow, an explicit direction--magnitude pair $(u, x)$ for Edge Flow. The second choice is how the oscillation amplitude is \emph{determined}: instantaneously, by projecting the gradient onto the tangent cone of $\{w \in \R^d : S(w) \le 2/\eta\}$ for Central Flow; dynamically, by its own differential equation, for Rod Flow and Edge Flow. In this appendix, we discuss in more details the connections and differences with Rod Flow, while Central Flow is abundantly discussed in the main paper.

Rod Flow \citep{regis2026rodflow} models the GD iterates as the endpoints of a ``rod''---a line segment in parameter space---whose center and orientation evolve according to a system of ODEs. The rod endpoint dynamics involve an auxiliary covariance-like matrix $\Sigma$ that governs the direction and magnitude of the oscillation and evolves (with our notation and time scaling) as
\begin{equation}    \label{eq:Sigma}
 \frac{d\Sigma}{dt} = \frac{\eta}{4} \big(\nabla L(\bar w_t + x_t u_t) \otimes \nabla L(\bar w_t + x_t u_t) + \nabla L(\bar w_t - x_t u_t) \otimes \nabla L(\bar w_t - x_t u_t)) - \frac{2}{\eta} \Sigma.   
\end{equation}
The direction $u_t$ and squared magnitude of the oscillation $x_t^2$ are then given by taking the first eigenpair of $\Sigma$. The first thing we can notice is that, arguably, the equations for Edge Flow are easier to interpret, since they make explicit the direction and magnitude of the oscillations instead of making them an implicit consequence of an eigendecomposition. This also makes Edge Flow less costly to implement and run.

\begin{figure}[t]
\centering
\includegraphics[width=0.66\textwidth]{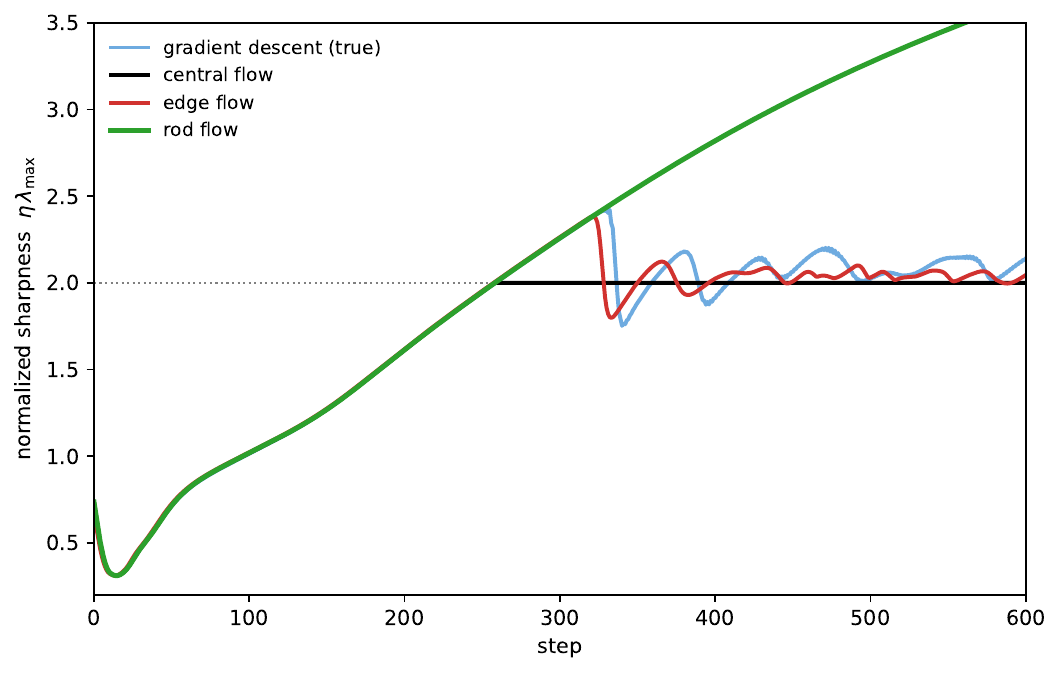}
\caption{\textbf{Sharpness stabilization by the three flows} ($3$-layer MLP, MSE loss on a $400$-example, $4$-class subset of CIFAR-10, $\eta = 0.017$). Normalized sharpness $\eta\lambda_{\max}$ at the center of each process: gradient descent (true dynamics), Central Flow, Edge Flow, and Rod Flow; the dotted line is the edge $\eta\lambda = 2$. Central Flow pins the sharpness at $2$ and Edge Flow tracks the gradient-descent oscillation around it, whereas Rod Flow does not stabilize the sharpness and drifts upward. Running Gradient Flow yields the same dynamics as Rod Flow.}
\label{fig:rod-sharpness}
\end{figure}

A more concrete problematic behavior of Rod Flow is that it fails to hold the sharpness at the edge of stability. Figure~\ref{fig:rod-sharpness} illustrates this by comparing the top effective-Hessian eigenvalue $\eta\lambda_{\max}$ at the center of each process along a training run, for gradient descent and the three flows. After the onset of EoS, Central Flow pins the sharpness exactly at $2$ (its defining constraint) and Edge Flow reproduces the overshoot and damped oscillation of gradient descent around $2$; Rod Flow, by contrast, lets the sharpness drift steadily upward, well past the threshold. 

We provide a heuristic interpretation for this phenomenon. Starting from \eqref{eq:Sigma}, assuming that the largest eigenvector of $\Sigma_t$ and the gradients at the rod endpoints are aligned with $u_t$, and doing a Taylor expansion of the gradients, we obtain that the magnitude of the oscillations evolves according to
\begin{align*}
    \frac{d(x_t^2)}{dt} \approx u_t^\top \frac{d\Sigma}{dt} u_t &\approx \frac{\eta}{4} (\|\nabla L(\bar w_t + x_t u_t)\|^2 + \|\nabla L(\bar w_t - x_t u_t)\|^2) - \frac{2}{\eta} x_t^2 \\
    &\approx  x_t^2 \Big(\frac{\eta}{2}\|\nabla^2 L(\bar w_t)\|^2 - \frac{2}{\eta}\Big) \\
    &\approx x_t^2 \Big(S(\bar w_t) - \frac{2}{\eta}\Big)
\end{align*}
where the last equation uses $S(\bar w_t) \approx 2/\eta$.
In terms of $x_t$, this reads $\frac{dx_t}{dt} \approx x_t (S(\bar w_t) - \frac{2}{\eta})$. We recover our evolution \eqref{eq:ef_x}, so at first sight Rod Flow and Edge Flow should behave similarly. However, a key assumption was made in this computation when going from the second step to the third: that the squared gradient norm at the \textit{current} oscillatory iterates is of the order of $\eta x_t^2 S(\bar w_t)^2 / 4$. This holds after the dynamics have stabilized at EoS, when the gradient at the endpoints is dominated by the oscillatory term which is indeed of this order. This is verified experimentally by \citet{regis2026rodflow}, which shows that Rod Flow dynamics track GD, when initialized well into the EoS phase. However, this does \textit{not} hold during progressive sharpening and at the onset of EoS, before the oscillations start their exponential growth, where the gradient is still dominated by $\nabla L(\bar w_t)$. Plugging this term in the second step above, we obtain that the Rod Flow dynamics then rather behaves like 
\begin{align*}
    \frac{d(x_t^2)}{dt} \approx \frac{\eta}{2} \|\nabla L(\bar w_t)\|^2  - \frac{2}{\eta} x_t^2.
\end{align*}
The key difference is that the expansive term in these dynamics does not depend on $x_t^2$, so it does not yield an exponential increase in $x_t$ when the sharpness crosses $2/\eta$. Experimentally the gradient norm at the center $\bar{w}_t$ is approximately constant in time around the onset of EoS, meaning that the equation above yields constant $x_t$ instead of exponentially increasing. This missing link between crossing the sharpness threshold and exponential increase in $x_t$ seems to be the reason why Rod Flow is not able to detect the onset of the EoS and to stabilize the sharpness.

\section{Experimental Details}
\label{app:experiments}

This appendix gives the details needed to reproduce the figures of Section~\ref{sec:experiments}. We use two settings: the \emph{main experiments}, spanning three architectures (a convolutional network, a residual network, and a vision transformer) and two losses (mean-squared error and cross-entropy) for the six configurations of Table~\ref{tab:exp-config}; and a \emph{simple MLP setting}, used to isolate the onset instabilities (Figures~\ref{fig:onset-base-level} and~\ref{fig:onset-lr}) and for the rod-flow comparison (Figure~\ref{fig:rod-sharpness}). The two settings share most of their setup, which we describe first.

\paragraph{Common setup.}
All experiments are implemented in \textbf{PyTorch}, building on the publicly available Central Flows codebase of \citet{cohen2025understanding} (\url{https://github.com/locuslab/central_flows}), which we extend to implement Edge Gradient Descent (see pseudocode in Appendix~\ref{app:experiments}); gradient descent and Central Flow processes use the original implementation. The data is a fixed subset of \textbf{CIFAR-10} restricted to the first $4$ classes (labels $0$--$3$), keeping the first $n/4$ of the $32\times 32\times 3$ images of each class with per-channel normalization. We train by full-batch gradient descent---the full batch of $n$ examples at every step, with no mini-batching, data augmentation, or weight decay---at a fixed learning rate $\eta$, in \texttt{float32} precision on a single GPU with random seed $0$. The flows are integrated with forward Euler, splitting each modeled gradient-descent step of size $\eta$ into $K = \eta/\rho$ substeps; Edge Flow starts from the same initialization as gradient descent, with the oscillation magnitude floored at a base level, $x \leftarrow \max(x, \varepsilon)$. Hessian eigenvalues and eigenvectors are computed from Hessian--vector products via automatic differentiation, so the full Hessian is never materialized.

\paragraph{Main experiments.}
The six configurations (Table~\ref{tab:exp-config}) combine three architectures with two losses---mean-squared error (MSE) regressed onto one-hot labels and softmax cross-entropy (CE)---and use $n=1000$ training examples ($250$ per class), with $4$ output units. Test metrics are evaluated on the CIFAR-10 test images of the same four classes, and the network-output panels display the model's four output logits on a single fixed test example. For each configuration we run gradient descent together with Edge Flow ($\rho = \eta/8$); for the CNN/MSE configuration we additionally run the Central Flow of \citet{cohen2025understanding} ($\rho = \eta/4$). The base level is set adaptively to $\varepsilon_t = |\nabla L(\bar w_t)^\top u_t|$ (see Section~\ref{sec:exp-setup}). Reported Hessian eigenvalues use a warm-started LOBPCG iteration (numerical tolerance $10^{-10}$), recomputed every step for the CNN and every $5$ steps for the ResNet and ViT (the eigensolve dominates the cost for the larger models). The three architectures, taken from \citet{cohen2025understanding}, with parameter counts for the $4$-class CIFAR-10 setting, are:
\begin{itemize}
    \item \textbf{CNN} ($\approx\!544$k parameters): two convolutional blocks with $32$ and $64$ channels respectively, followed by a bias-free linear layer to a hidden width of $128$, a GELU, and a linear read-out. Width hyperparameter $w=32$.
    \item \textbf{ResNet} ($\approx\!293$k parameters): a pre-projection convolution followed by three stages of three residual blocks each (two $3\times3$ convolutions per block with GroupNorm and GELU), with spatial downsampling and channel doubling between stages; the resulting $8\times8$ feature map is reduced by a $4\times4$ average-pooling layer to a $2\times2$ map, flattened, and passed to a linear read-out. Base width $w=16$. Normalization is GroupNorm.
    \item \textbf{ViT} ($\approx\!661$k parameters): the \textsc{SimpleViT} architecture from the \texttt{vit-pytorch} library with patch size $4$, embedding dimension $64$, depth $4$, $8$ attention heads, and MLP dimension $256$.
\end{itemize}

\begin{table}[t]
\centering
\caption{Configurations of the six experiments. All use full-batch gradient descent on a $1000$-example, $4$-class subset of CIFAR-10. ``Processes'' lists the trajectories run in addition to gradient descent (GD): Edge Flow (EF) and, where applicable, Central Flow (CF). The Central Flow is run only for the CNN/MSE configuration.}
\label{tab:exp-config}
\begin{tabular}{llccccc}
\toprule
Architecture & \#\,params & Loss & $\eta$ & Steps & Processes & Eig.\ frequency \\
\midrule
CNN    & $544$k & MSE & $0.02$ & $2000$ & GD, EF, CF & $1$ \\
CNN    & $544$k & CE  & $0.01$ & $3000$ & GD, EF     & $1$ \\
ResNet & $293$k & MSE & $0.02$ & $3500$ & GD, EF     & $5$ \\
ResNet & $293$k & CE  & $0.02$ & $3500$ & GD, EF     & $5$ \\
ViT    & $661$k & MSE & $0.02$ & $3000$ & GD, EF     & $5$ \\
ViT    & $661$k & CE  & $0.02$ & $4500$ & GD, EF     & $5$ \\
\bottomrule
\end{tabular}
\end{table}

\paragraph{Metrics.} On top of the metrics already discussed and reported in the main text (see Figure~\ref{fig:cnn-mse-overlay}), we report the following additional metrics:
\begin{itemize}
    \item \textbf{Oscillation variance.} Edge Flow predicts the variance of the iterate around its center to be $x_t^2$. This is compared against the empirical GD oscillation variance, estimated by the squared half-step of gradient descent $\tfrac{\eta^2}{4}\|\nabla L(w_t)\|^2$. Note that the variance of the oscillations can also be read off through the relation 
    $  x_t u_t \approx \tfrac{\eta}{4}(g_+ - g_-)$.
     This equation comes from a self-consistency observation for the oscillation $x_t u_t$: it should align with the gradient at the two rod endpoints. This yields the estimate $\tfrac{\eta^2}{16}\|g_+ - g_-\|^2$ of $x_t^2$. For this reason, after the EoS onset, the oscillation variance is essentially proportional to the squared gradient norm, as can be observed in our plots.
     \item \textbf{Network outputs on a fixed test example.} We report the evolution during training time of the output of the network for a fixed input. The output is four-dimensional, for each training method we report each of the four dimensions as a univariate curve. The shade encodes the output index.
\end{itemize}

\paragraph{Simple MLP setting.}
The onset-instability figures (Figures~\ref{fig:onset-base-level} and~\ref{fig:onset-lr}) and the rod-flow comparison (Figure~\ref{fig:rod-sharpness}) use a smaller, self-contained instance of the setup above. Relative to the main experiments, three things change: the architecture is a $3$-layer multilayer perceptron with GELU activations (layer sizes $3072 \to 64 \to 64 \to 4$; hidden width $16$ for the Rod Flow comparison); the data is the smaller $n=400$ subset ($100$ images per class); and only the MSE loss is used. Edge Gradient Descent is run as in Definition~\ref{def:egd}, with $K$ substeps of size $\rho = \eta/K$ and the always-on floor $x \leftarrow \max(x, \varepsilon)$ (default base level $\varepsilon = 10^{-5}$, swept in Figure~\ref{fig:onset-base-level}). The Edge Flow sharpness is reported as the power-iteration estimate $\|\nabla^2 L(\bar w)\,u\|$ (we checked on some of the experiments that it gives similar results to warm-started LOBPCG for a much lower cost). The per-figure settings are:
\begin{itemize}
    \item \textbf{Figure~\ref{fig:onset-base-level} (base level):} $\eta = 0.02$, $1000$ steps, two discretizations $K \in \{1, 8\}$, sweeping the base level $\varepsilon \in \{10^{-9}, 10^{-7}, 10^{-5}, 10^{-3}\}$. The gradient-descent oscillation magnitude is estimated by $\tfrac{\eta}{2}\|\nabla L\|$.
    \item \textbf{Figure~\ref{fig:onset-lr} (learning rate):} $1000$ steps, two discretizations $K \in \{1, 4\}$, fixed $\varepsilon = 10^{-5}$, sweeping $\eta \in \{0.05, 0.1, 0.2, 0.4\}$; the horizontal axis is the flow time $\eta t$, the light trace is the raw per-step sharpness and the bold curve a moving-average trend, and the gradient-descent oscillation magnitude is estimated by $\tfrac{\eta}{2}\|\nabla L\|$.
    \item \textbf{Figure~\ref{fig:rod-sharpness} (rod-flow comparison):} hidden width $16$, $\eta = 0.017$, $800$ steps, top $3$ Hessian eigenvalues by warm-started LOBPCG; $K = 8$ for Edge and Rod Flow, $K = 4$ for Central Flow. The figure plots the top effective eigenvalue $\eta\lambda_{\max}$ at the center of each process.
\end{itemize}

\paragraph{Pseudocode for Edge Gradient Descent.}
Algorithm~\ref{alg:egd} gives the practical form of Edge Gradient Descent (EGD) used in our experiments. It is the forward-Euler/power-method discretization of Edge Flow introduced in Section~\ref{sec:egd}, augmented with the two implementation details discussed in the main text: each modeled gradient-descent step of size $\eta$ is split into $K$ substeps of size $\rho = \eta/K$, and the oscillation magnitude is floored at the base level $\varepsilon$.
Simulating $T$ gradient-descent steps therefore costs $T\,K$ substeps, each requiring two gradient evaluations and one Hessian--vector product.

\begin{algorithm}[t]
\caption{Edge Gradient Descent (EGD)}
\label{alg:egd}
\begin{algorithmic}[1]
\Require loss $L$, learning rate $\eta$, substeps per step $K$ (so $\rho = \eta/K$), base level $\varepsilon > 0$, number of modeled GD steps $T$, initialization $w_0$
\State $\bar w \gets w_0$ \Comment{center}
\State $u \gets$ top eigenvector of $\nabla^2 L(\bar w)$ \Comment{e.g.\ a few power iterations; $\|u\| = 1$}
\State $x \gets \varepsilon$ \Comment{oscillation magnitude}
\For{$t = 0, 1, \dots, T-1$}
    \For{$j = 1, \dots, K$}
        \State $g \gets \tfrac{1}{2}\big(\nabla L(\bar w + x\,u) + \nabla L(\bar w - x\,u)\big)$ \Comment{symmetrized endpoint gradient}
        \State $v \gets \nabla^2 L(\bar w)\,u$ \Comment{Hessian--vector product (one autodiff pass)}
        \State $S \gets \|v\|$ \Comment{sharpness estimate $S(\bar w)$}
        \State $\bar w \gets \bar w - \rho\, g$ \Comment{center update \eqref{eq:egd_w}}
        \State $u \gets v / \|v\|$ \Comment{direction: one power iteration \eqref{eq:egd_u}}
        \State $x \gets x\,\big(1 + \rho\,(S - 2/\eta)\big)$ \Comment{magnitude update \eqref{eq:egd_x}}
        \State $x \gets \max(x, \varepsilon)$ \Comment{floor at the base level}
    \EndFor
\EndFor
\State \Return center trajectory $(\bar w)$
\end{algorithmic}
\end{algorithm}

\section{Additional Experimental Figures}
\label{app:additional-figures}

This appendix collects additional figures, first for the main experiments and then for the simple MLP setting.

\subsection{Main experiments}

\begin{figure}[t]
\centering
\includegraphics[width=\textwidth]{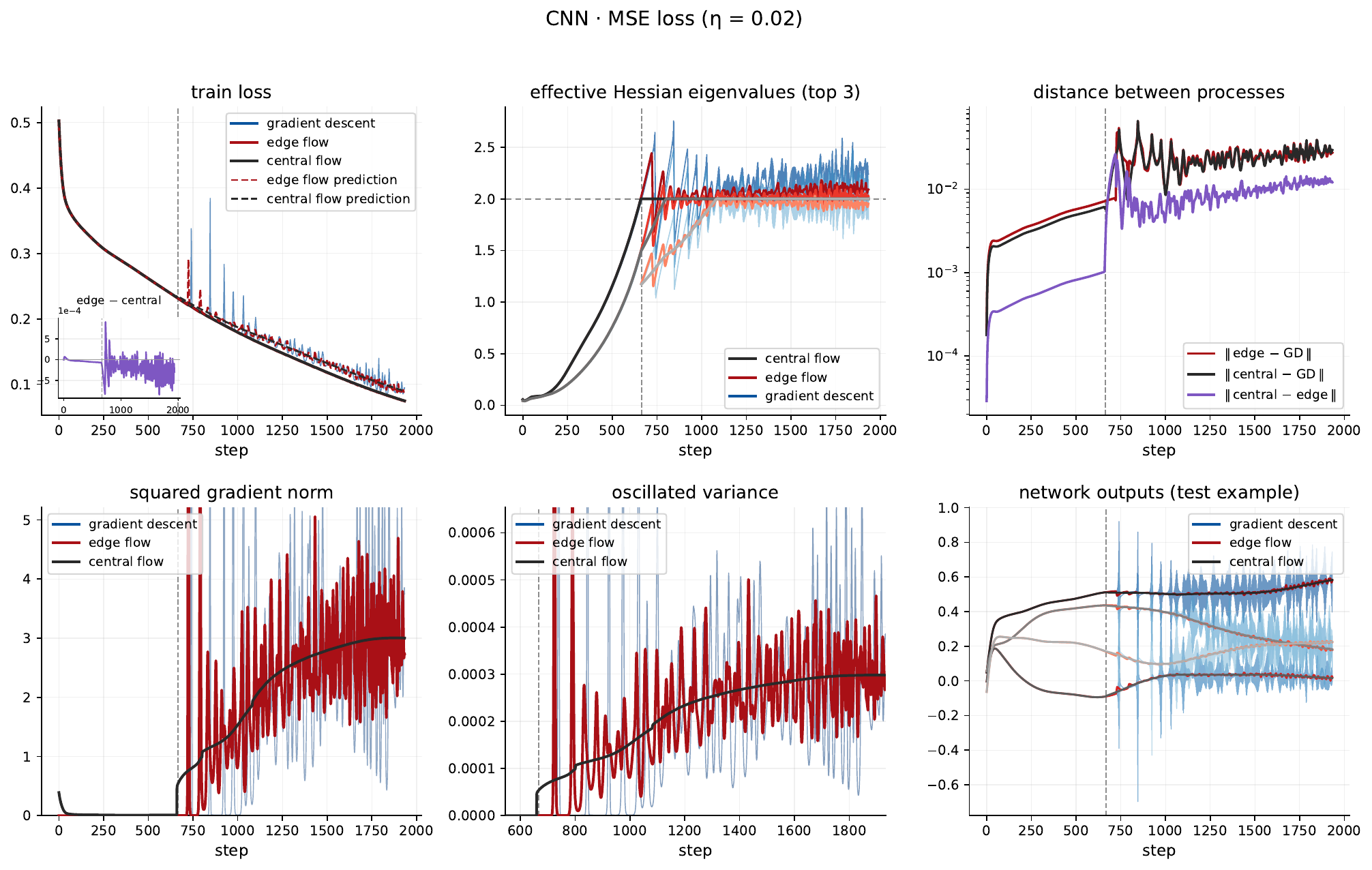}\\[6pt]
\includegraphics[width=0.72\textwidth]{figures/cnn-mse_eos.pdf}
\caption{\textbf{CNN with MSE loss} ($\eta = 0.02$): full set of metrics (top) and zoom on the onset of the edge of stability (bottom). The inset on the train losses panels shows the small gap between the Edge Flow and Central Flow centers. Some of these figures are also in the main text. }
\label{fig:cnn-mse}
\end{figure}

\begin{figure}[t]
\centering
\includegraphics[width=\textwidth]{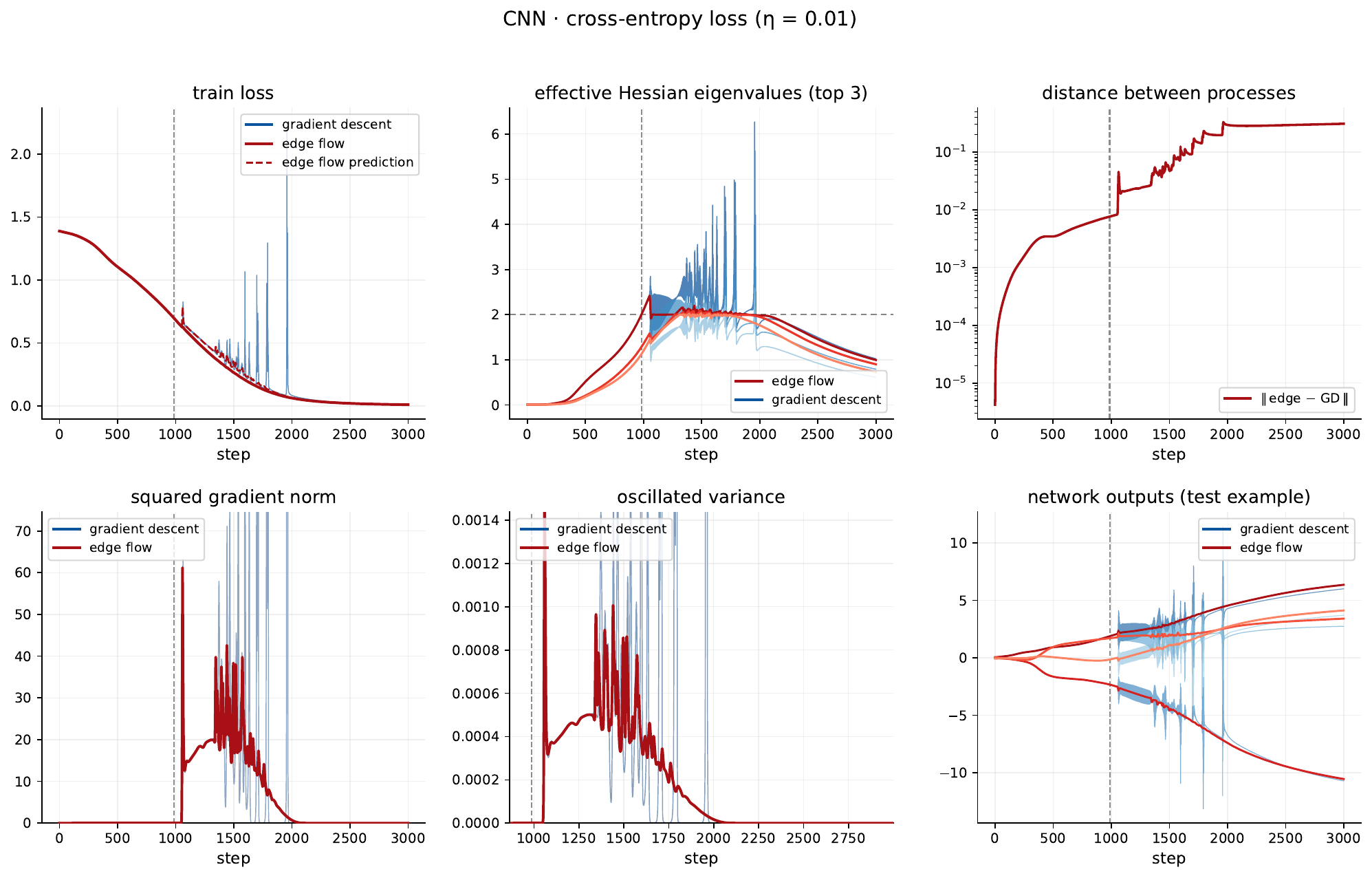}\\[6pt]
\includegraphics[width=0.72\textwidth]{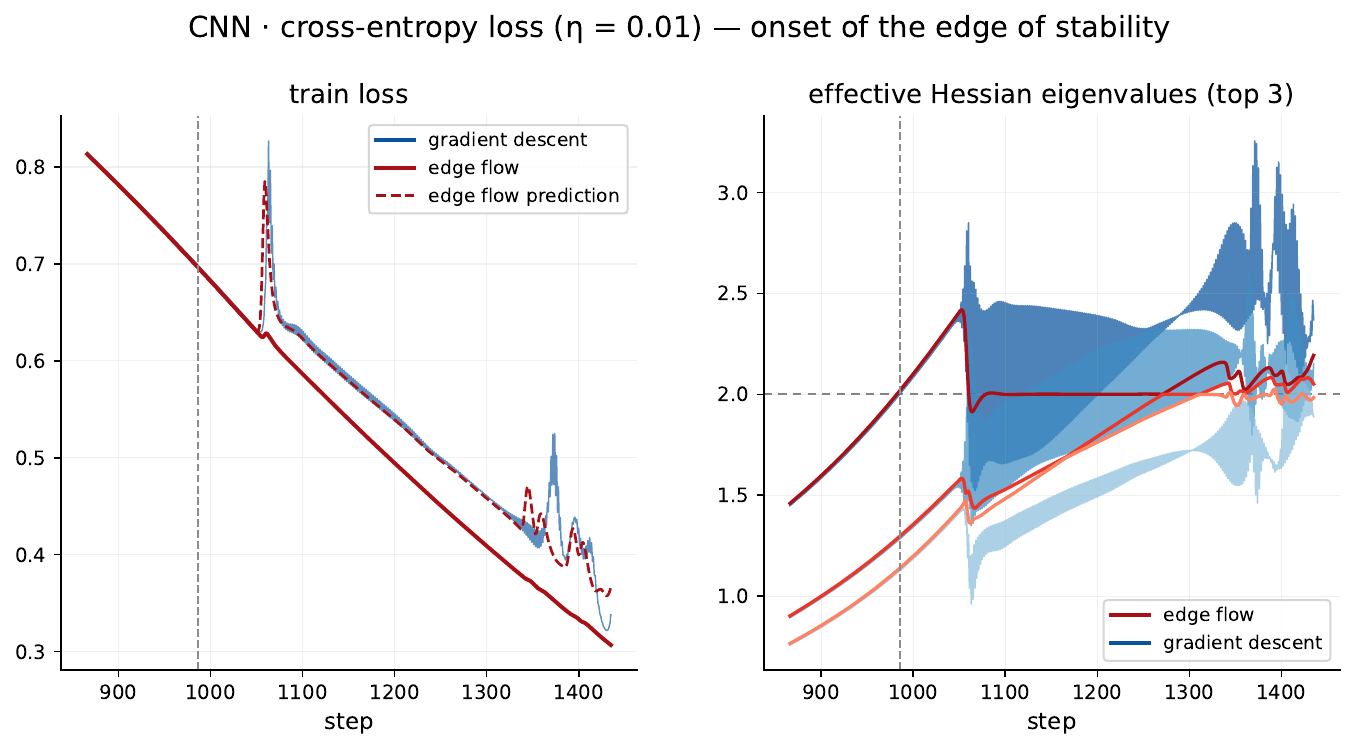}
\caption{\textbf{CNN with cross-entropy loss} ($\eta = 0.01$). Full set of metrics (top) and zoom on the onset of the edge of stability (bottom).}
\label{fig:cnn-ce}
\end{figure}

\begin{figure}[t]
\centering
\includegraphics[width=\textwidth]{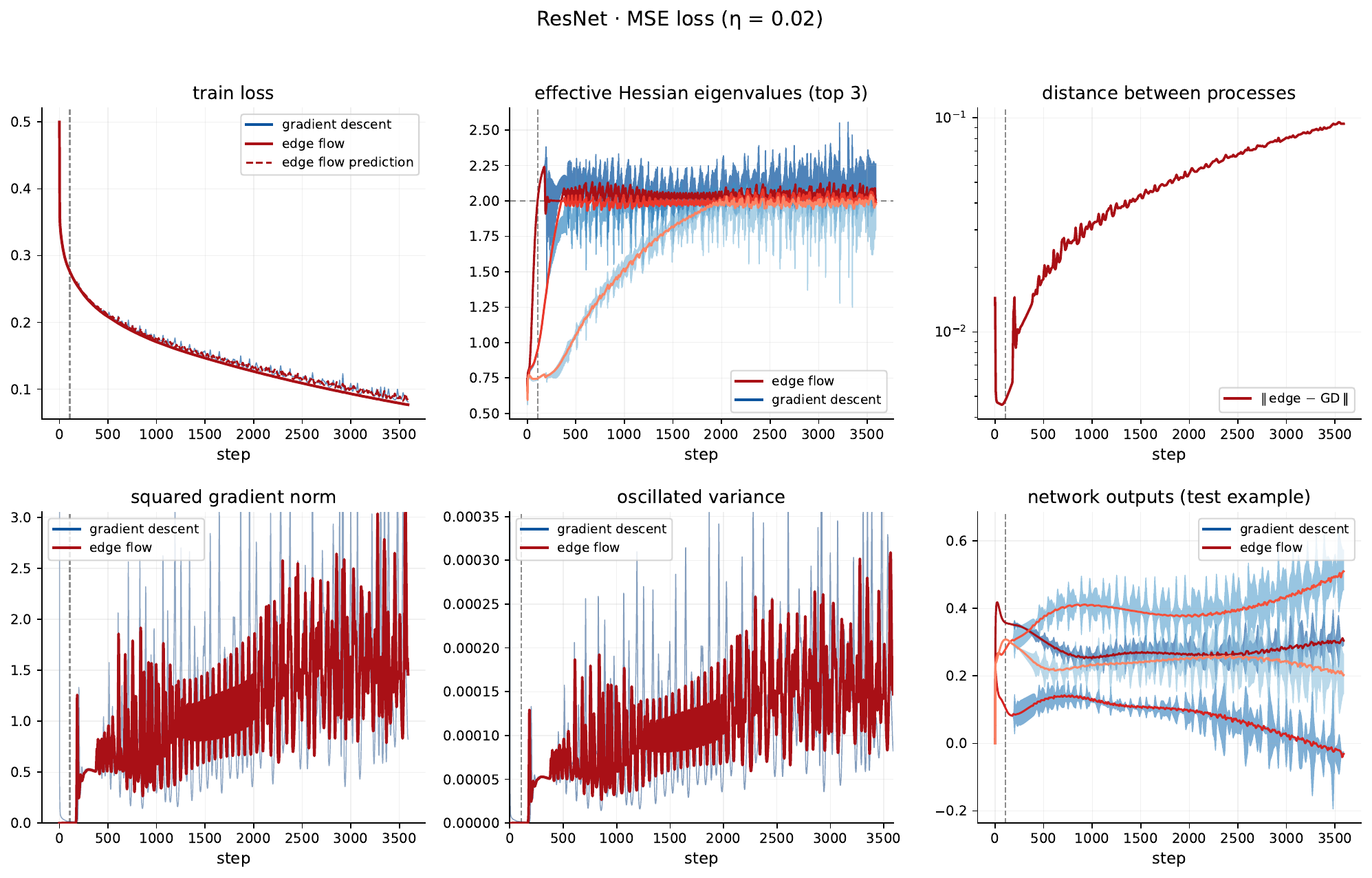}\\[6pt]
\includegraphics[width=0.72\textwidth]{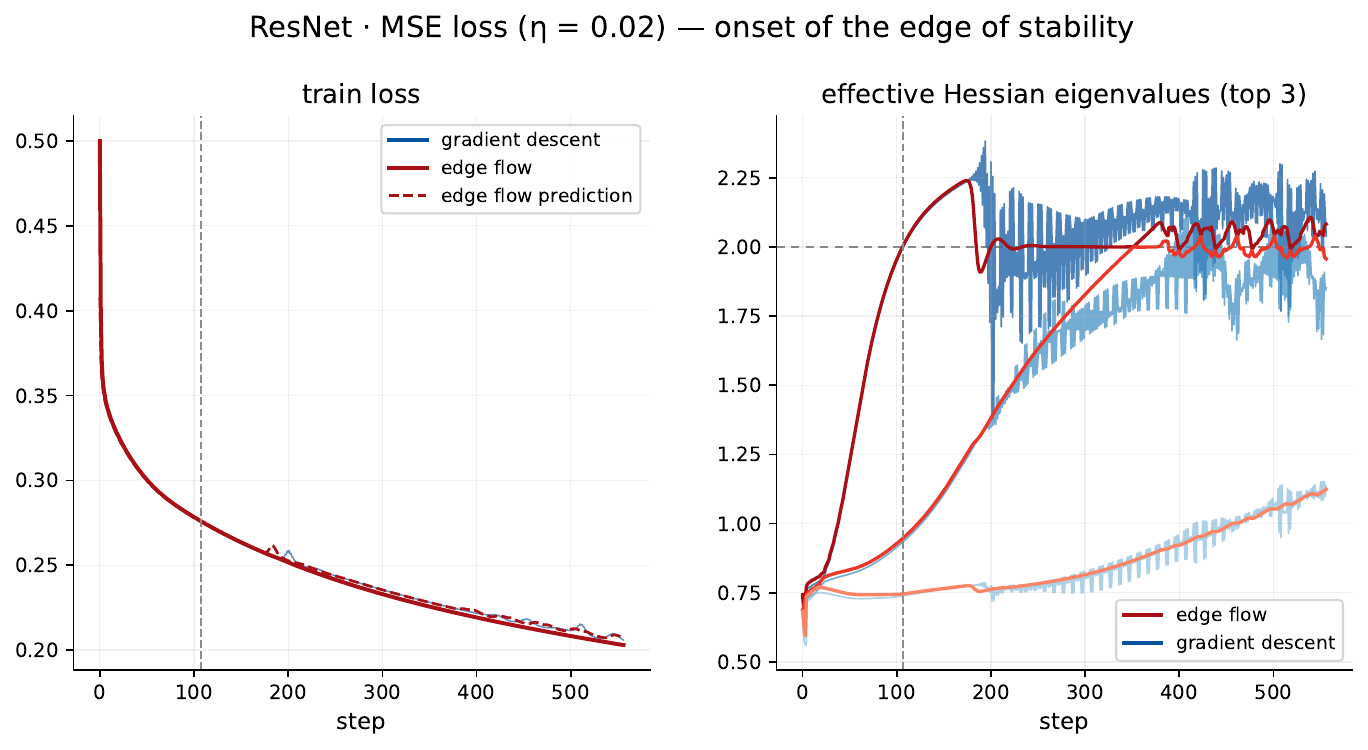}
\caption{\textbf{ResNet with MSE loss} ($\eta = 0.02$). Full set of metrics (top) and zoom on the onset of the edge of stability (bottom).}
\label{fig:resnet-mse}
\end{figure}

\begin{figure}[t]
\centering
\includegraphics[width=\textwidth]{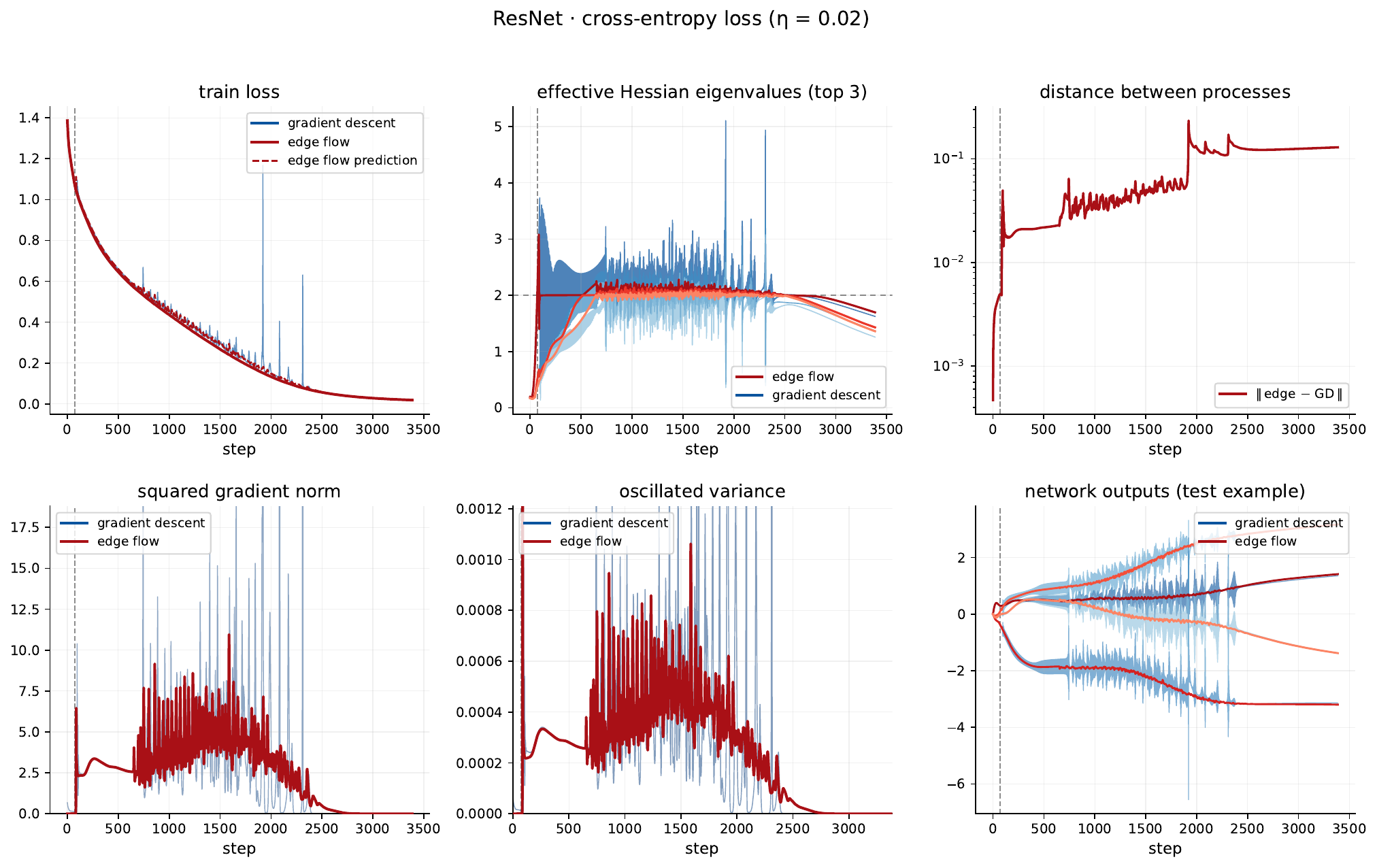}\\[6pt]
\includegraphics[width=0.72\textwidth]{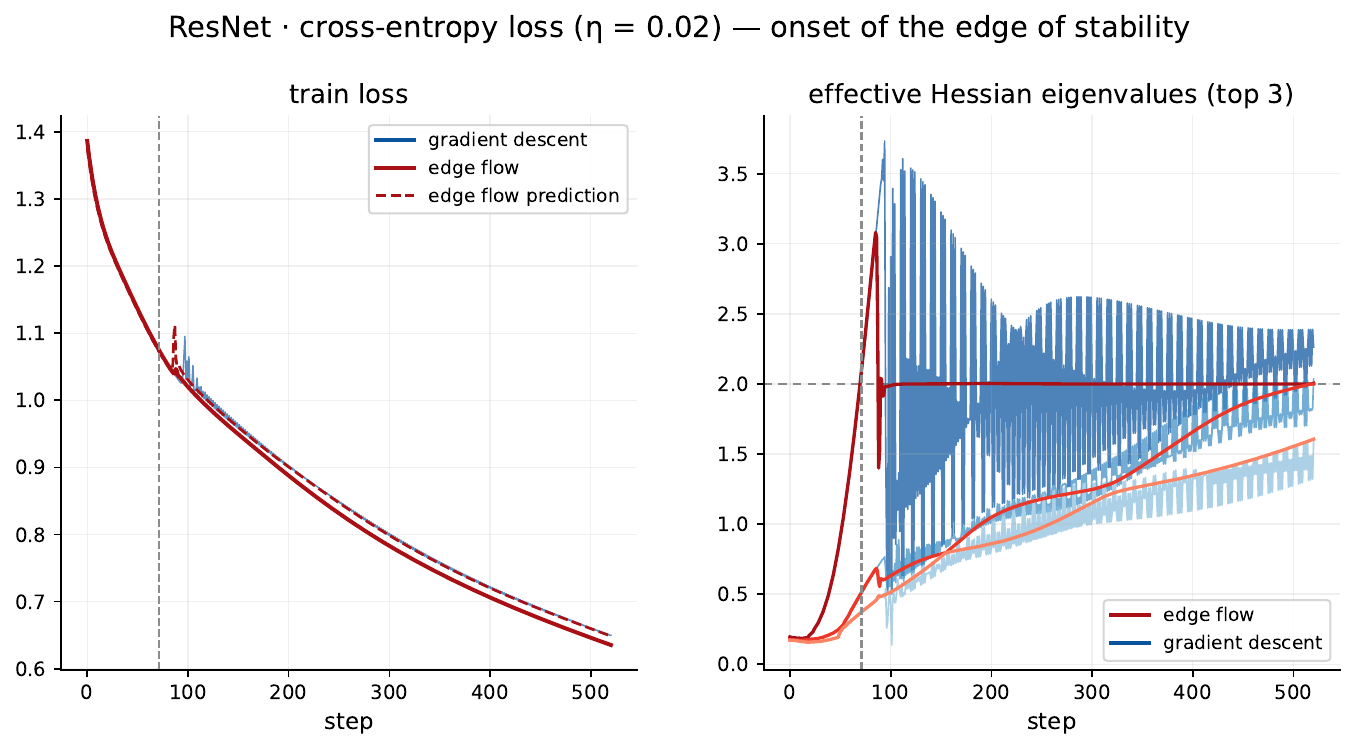}
\caption{\textbf{ResNet with cross-entropy loss} ($\eta = 0.02$). Full set of metrics (top) and zoom on the onset of the edge of stability (bottom).}
\label{fig:resnet-ce}
\end{figure}

\begin{figure}[t]
\centering
\includegraphics[width=\textwidth]{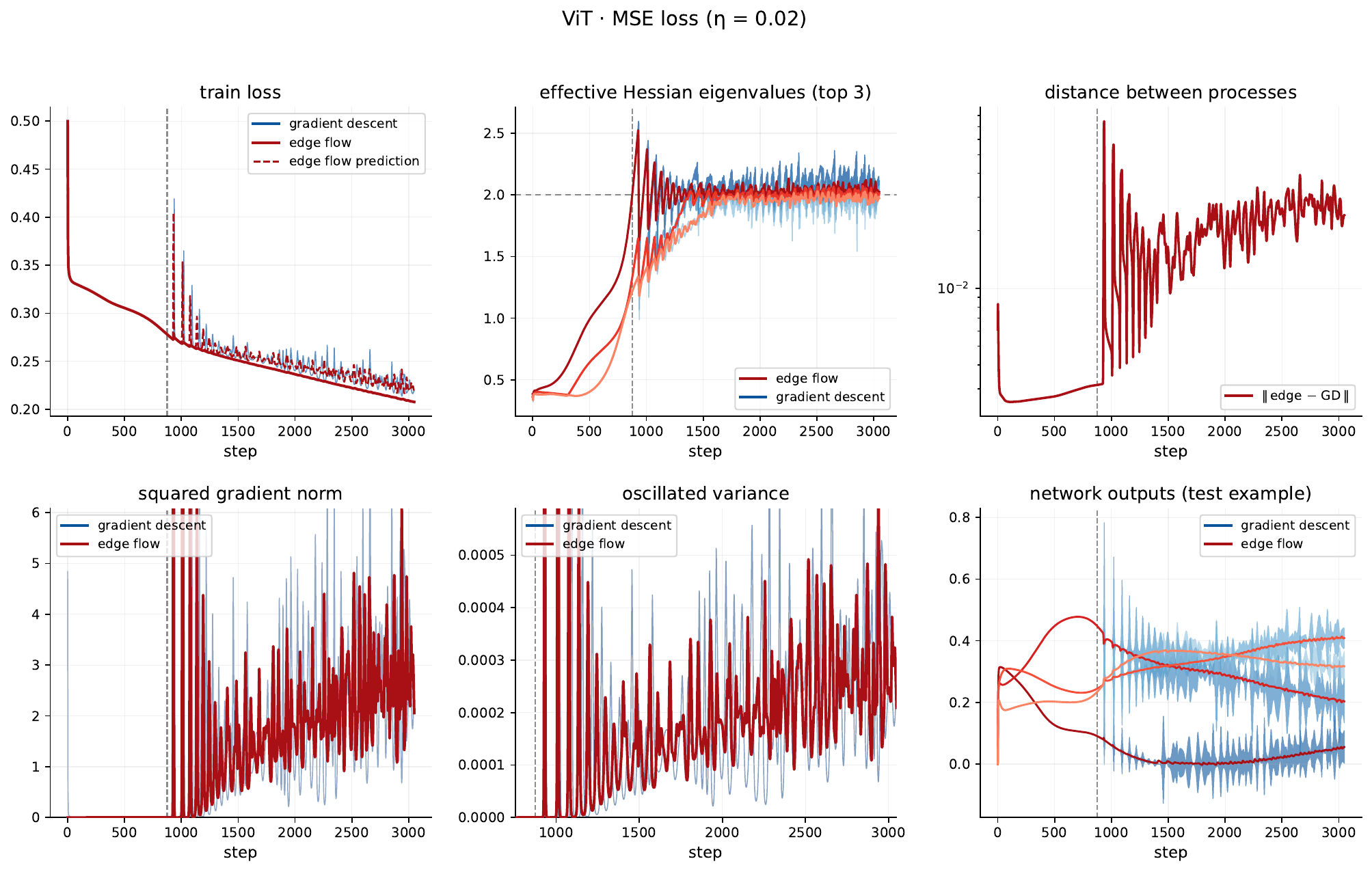}\\[6pt]
\includegraphics[width=0.72\textwidth]{figures/vit-mse_eos.pdf}
\caption{\textbf{ViT with MSE loss} ($\eta = 0.02$): full set of metrics and zoom on the onset of the edge of stability (bottom). Some of these figures are also in the main text.}
\label{fig:vit-mse}
\end{figure}

\begin{figure}[t]
\centering
\includegraphics[width=\textwidth]{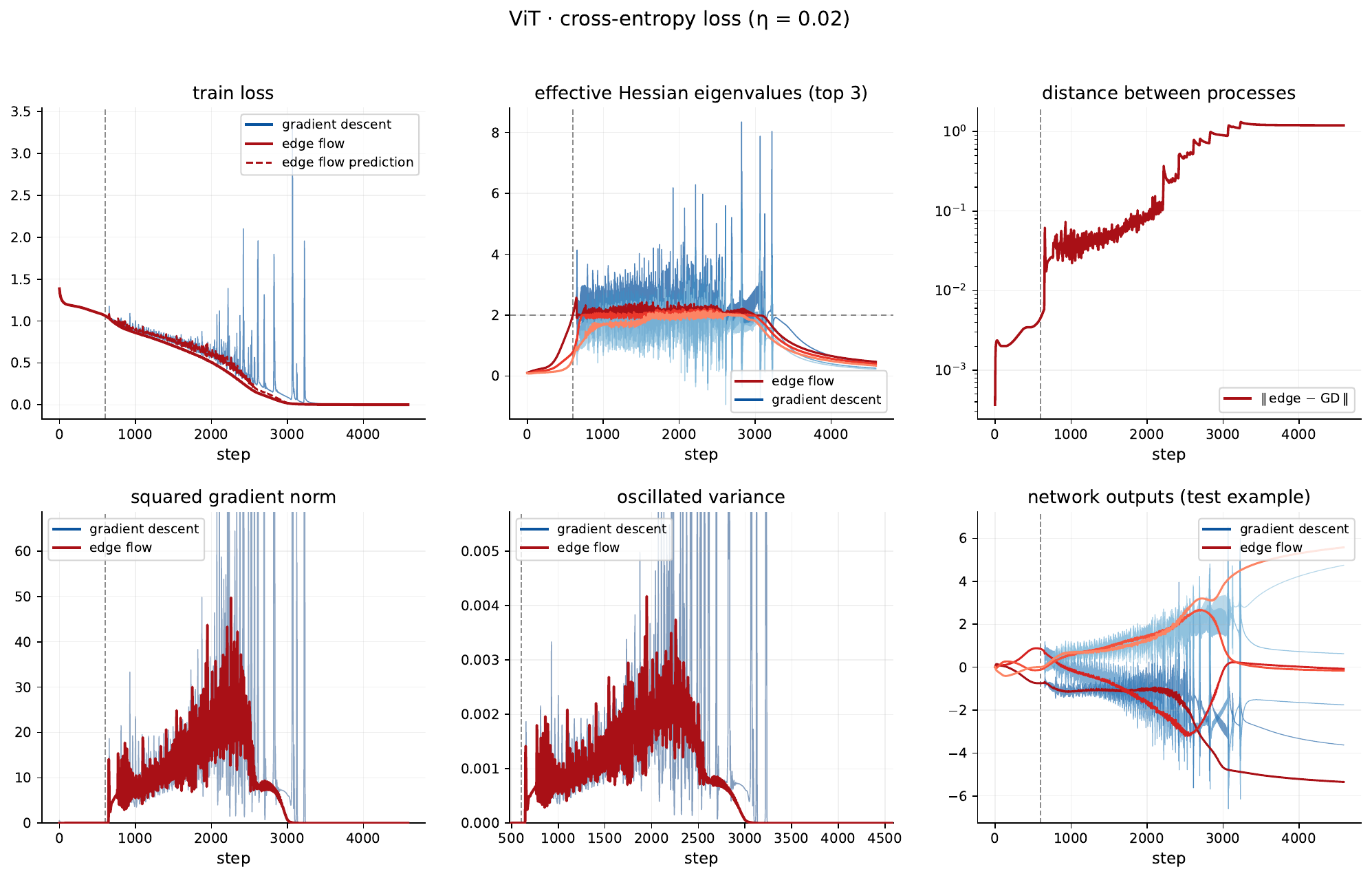}\\[6pt]
\includegraphics[width=0.72\textwidth]{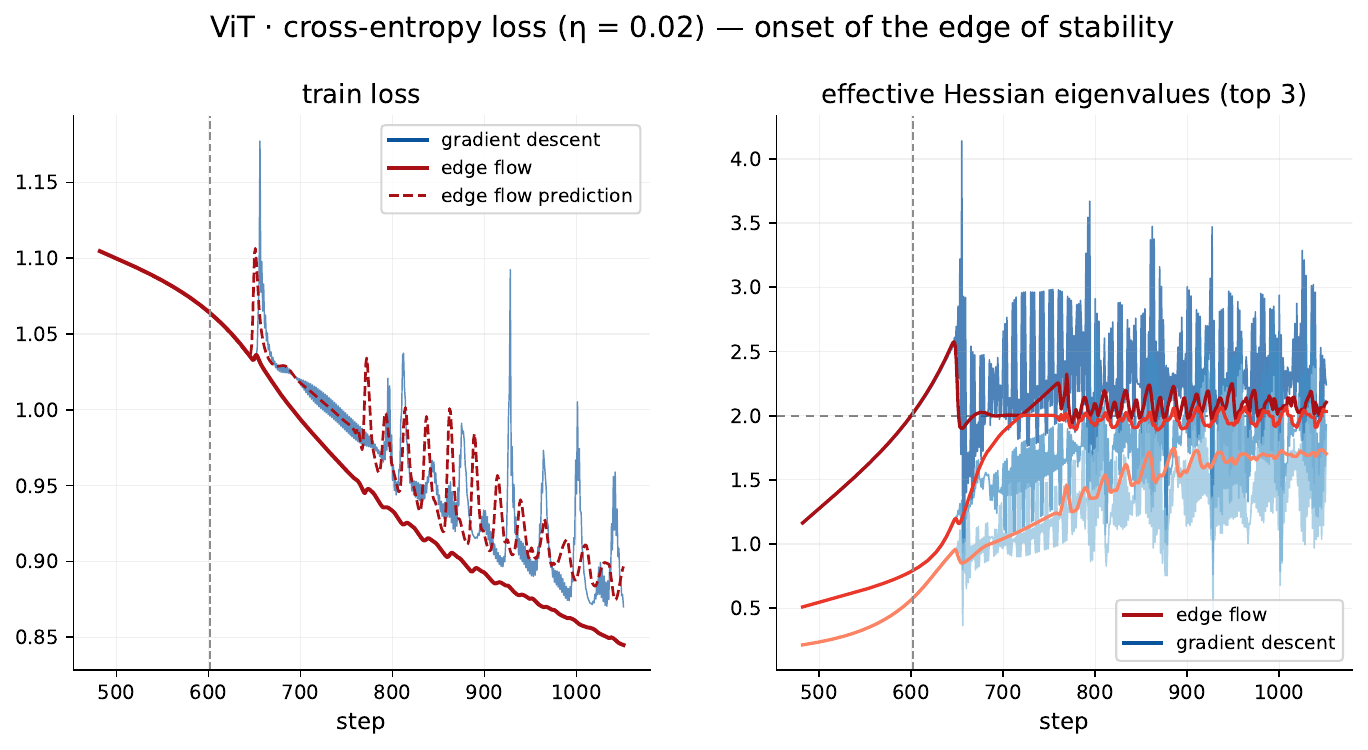}
\caption{\textbf{ViT with cross-entropy loss} ($\eta = 0.02$). Full set of metrics (top) and zoom on the onset of the edge of stability (bottom).}
\label{fig:vit-ce}
\end{figure}

We report results for the six architecture/loss configurations of Table~\ref{tab:exp-config} in Figures~\ref{fig:cnn-mse} to~\ref{fig:vit-ce}. For each configuration we show the full set of metrics, and a zoom on the onset of the edge of stability. In every case the Edge Flow tracks the gradient-descent trajectory across all metrics and reproduces the overshoot and damped oscillation of the sharpness at the onset of EoS, as well as the exit times of EoS in the case of cross-entropy loss.

\subsection{Simple MLP setting: effect of the learning rate}
\label{app:onset}

\begin{figure}[t]
\centering
\includegraphics[width=\textwidth]{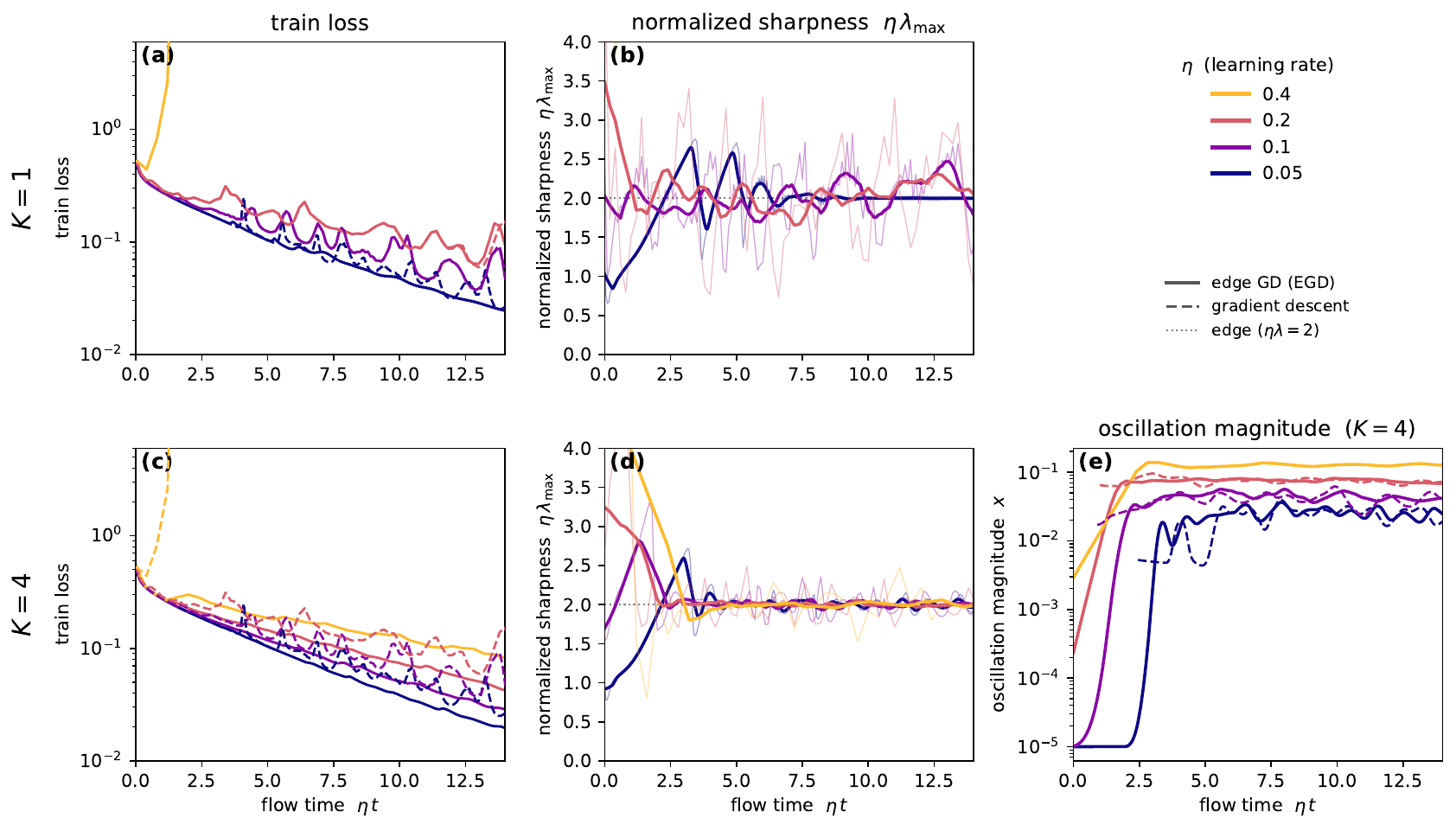}
\caption{\textbf{Effect of the learning rate on stability}, for the MLP/MSE setup of Figure~\ref{fig:onset-base-level}, at a coarse ($K=1$, top) and a fine ($K=4$, bottom) discretization. Colour encodes the learning rate $\eta$ and the horizontal axis is the flow time $\eta t$. \emph{Columns:} training loss (with the matched gradient-descent reference, dashed); normalized sharpness $\eta\lambda_{\max}$ (light: raw per-step value; bold: smoothed trend; dotted: edge $=2$); and, for $K=4$, the oscillation magnitude $x$. At the largest learning rate ($\eta = 0.4$) the $K=1$ Edge Flow and gradient descent diverge, whereas $K=4$ remains stable: finer discretization of the center dynamics extends the range of stable learning rates. Larger $K$ also reduces the sharpness oscillations.}
\label{fig:onset-lr}
\end{figure}

Figure~\ref{fig:onset-lr} complements Figure~\ref{fig:onset-base-level} by varying the learning rate $\eta$ at two discretizations, $K=1$ and $K=4$, for the same MLP/MSE setup. The horizontal axis is the flow time $\eta t$, on which the gradient flow trajectory is $\eta$-independent; accordingly the training loss curves for different $\eta$ overlap during the initial descent and separate only once each enters the edge of stability---earlier, in flow time, for larger $\eta$.

A key observation is the contrast between the two rows. At the largest learning rate, $\eta = 0.4$, the coarse $K=1$ scheme (and the gradient-descent reference) diverge within a handful of steps, whereas the fine $K=4$ scheme continues to train stably, its sharpness damping onto the edge $\eta\lambda = 2$. Pushing further, $K=4$ also eventually diverges, around $\eta = 0.5$. Refining the discretization of the center thus seems to extend the range of stable learning rates, as discussed in Section~\ref{sec:instabilities}.

\end{document}